%% file: main.tex
\documentclass[10pt, twocolumn, a4paper]{article}

\input{utils/include}

\input{utils/general_utils}

\input{utils/math_utils}

\usepackage[numbers,square]{natbib}
\usepackage[utf8]{inputenc}	
\usepackage[T1]{fontenc}	
\usepackage{xcolor}		
\usepackage[colorlinks = true,
            linkcolor = purple,
            urlcolor  = blue,
            citecolor = cyan,
            anchorcolor = black]{hyperref}	
\usepackage{booktabs} 		
\usepackage{nicefrac}		
\usepackage{microtype}		
\usepackage{lineno}		
\usepackage{float}			

\usepackage{lipsum}		

\usepackage{titling}
\usepackage{newfloat}
\DeclareFloatingEnvironment[name={Supplementary Figure}]{suppfigure}
\usepackage{sidecap}
\sidecaptionvpos{figure}{c}

\usepackage{titlesec}
\titlespacing\section{0pt}{12pt plus 3pt minus 3pt}{1pt plus 1pt minus 1pt}
\titlespacing\subsection{0pt}{10pt plus 3pt minus 3pt}{1pt plus 1pt minus 1pt}
\titlespacing\subsubsection{0pt}{8pt plus 3pt minus 3pt}{1pt plus 1pt minus 1pt}

\usepackage{tikz,xcolor,hyperref}
\usepackage{cleveref}

\definecolor{lime}{HTML}{A6CE39}
\DeclareRobustCommand{\orcidicon}{
	\begin{tikzpicture}
	\draw[lime, fill=lime] (0,0) 
	circle [radius=0.16] 
	node[white] {{\fontfamily{qag}\selectfont \tiny ID}};
	\draw[white, fill=white] (-0.0625,0.095) 
	circle [radius=0.007];
	\end{tikzpicture}
	\hspace{-2mm}
}
\foreach \x in {A, ..., Z}{\expandafter\xdef\csname orcid\x\endcsname{\noexpand\href{https://orcid.org/\csname orcidauthor\x\endcsname}
			{\noexpand\orcidicon}}
}

\usepackage{authblk}

\author[1,2,*]{Peter Lorenz}
\author[1]{Ricard Durall}
\author[3]{Janis Keuper}
\affil[1]{Fraunhofer ITWM}
\affil[2]{Heidelberg University}
\affil[3]{IMLA, Offenburg University and University of Mannheim}
\affil[*]{Correspondence to peter.lorenz@itwm.fhg.de}

\input{content/head}

\input{utils/acronym}

\begin{document}

\date{\vspace{-0ex}}

\twocolumn[ 
  \begin{@twocolumnfalse} 
\maketitle


  \end{@twocolumnfalse} 
] 


\begin{abstract}
\input{content/abstract}

\end{abstract}


\input{content/content.tex}



\section*{Acknowledgment}
We express our gratitude to Prof. Ullrich Köthe for diligently proofreading this paper.

\FloatBarrier
\small{ 

}



\appendix
\normalsize{
\include{content/supplementary}
}


\end{document}

%% file: utils/include.tex
\pdfoutput=1
\usepackage[pdftex]{graphicx}

\usepackage{multirow,mathtools} 
\usepackage{algorithm}
\usepackage{algpseudocode}
\usepackage{algorithmicx}

\usepackage{pifont}
\usepackage{color, colortbl}

\usepackage{blindtext}
\usepackage{lipsum}
\usepackage{threeparttable}

\usepackage{amssymb}
\usepackage{wrapfig}
\usepackage{times}
\usepackage{multirow}
\usepackage{url}
\usepackage{subfigure,epsfig,epstopdf}

\usepackage{color}
\usepackage{xcolor}

\usepackage{enumitem}
\usepackage{xspace}
\usepackage{xfrac}
\usepackage{comment}
\usepackage{tikz}
\usepackage{etoolbox}

\usepackage{placeins}
\usepackage{subcaption}

\definecolor{SL_color}{rgb}{0.858, 0.188, 0.478}

\usepackage{pifont}
\usepackage{adjustbox}
\usepackage{threeparttable,booktabs}
\usepackage{multirow}

\DeclareMathAlphabet\mathbfcal{OMS}{cmsy}{b}{n}

\usepackage{amsmath}
\usepackage{MnSymbol}
\usepackage{wasysym}

\usepackage{mathtools}
\DeclarePairedDelimiterX{\inp}[2]{\langle}{\rangle}{#1, #2}

\usepackage{acro}

\usepackage{color, colortbl}
\definecolor{Gray}{gray}{0.93}
\definecolor{Orange}{rgb}{1,0.5,0}
\definecolor{DGray}{gray}{0.83}
\definecolor{LightCyan}{rgb}{0.88,1,1}

\RequirePackage[normalem]{ulem} 
\RequirePackage{color}\definecolor{RED}{rgb}{1,0,0}\definecolor{BLUE}{rgb}{0,0,1} 


%% file: utils/general_utils.tex
\usepackage{algorithm,algpseudocode}
\algnewcommand{\algorithmicforeach}{\textbf{for each}}
\algdef{SE}[FOR]{ForEach}{EndForEach}[1]
  {\algorithmicforeach\ #1\ \algorithmicdo}
  {\algorithmicend\ \algorithmicforeach}

\usepackage{pifont}

\usepackage{color, colortbl}
\definecolor{Gray}{gray}{0.93}
\definecolor{Orange}{rgb}{1,0.5,0}
\definecolor{DGray}{gray}{0.83}
\definecolor{LightCyan}{rgb}{0.88,1,1}

\usepackage[most]{tcolorbox}

%% file: utils/math_utils.tex
\usepackage{amsmath,amsfonts,bm}

\def\eqref#1{(\ref{#1})}

\def\1{\bm{1}}

\def\vx{{\bm{x}}}
\def\vy{{\bm{y}}}

\def\mJ{{\bm{J}}}

\DeclareMathAlphabet{\mathsfit}{\encodingdefault}{\sfdefault}{m}{sl}
\SetMathAlphabet{\mathsfit}{bold}{\encodingdefault}{\sfdefault}{bx}{n}



%% file: content/head.tex
\title{Adversarial Examples are Misaligned in Diffusion Model Manifolds}

\hypersetup{
pdftitle={Manifold Mismatch},
pdfsubject={cs.CV},
pdfauthor={peter lorenz},
pdfkeywords={dadversarial examples, manifold, mismatch},
}

%% file: utils/acronym.tex
\usepackage{acro}

\DeclareAcronym{knn}{
  short=k-nn,
  long=k-nearest neighbor,
}

\DeclareAcronym{nnif}{
  short=NNIF,
  long=nearest neighbor and influnce functions,
}

\DeclareAcronym{wrn}{
  short=WRN,
  long=wide residual networks,
}

\DeclareAcronym{cnn}{
  short=CNN,
  long=convolutional neural networks,
}

\DeclareAcronym{at}{
  short=AT,
  long=adversarial training,
}

\DeclareAcronym{pca}{
  short=PCA,
  long=principal component analysis,
}

\DeclareAcronym{fnr}{
  short=FNR,
  long=false negative rate,
}

\DeclareAcronym{asr}{
  short=ASR,
  long=adversarial succes rate,
}

\DeclareAcronym{asrd}{
  short=ASRD,
  long=adversarial success rate under detection,
}

\DeclareAcronym{bb}{
  short=BB,
  long=black-box,
}

\DeclareAcronym{wb}{
  short=WB,
  long=white-box,
}

\DeclareAcronym{kd}{
  short=KD,
  long=kernel density,
}

\DeclareAcronym{id}{
  short=ID,
  long=intrinsic dimensionality,
}

\DeclareAcronym{lid}{
  short=LID,
  long=local intrinsic dimensionality,
}

\DeclareAcronym{mah}{
  short=M-D,
  long=mahalanobis distance,
}

\DeclareAcronym{sota}{
  short=SOTA,
  long=state-of-the-art,
}

\DeclareAcronym{dft}{
  short=DFT,
  long=discrete Fourier transformation,
}

\DeclareAcronym{fft}{
  short=FFT,
  long=Fast Fourier Transformation,
}

\DeclareAcronym{mfs}{
  short=MFS,
  long=magnitude Fourier spectrum,
}

\DeclareAcronym{pfs}{
  short=PFS,
  long=phase Fourier spectrum,
}

\DeclareAcronym{dnn}{
  short=DNN,
  long=Deep Neural Network,
}

\DeclareAcronym{fgsm} {
  short=FGSM,
  long=Fast Gradient Method,
}

\DeclareAcronym{bim} {
  short=BIM,
  long=Basic Iterative Method,
}

\DeclareAcronym{autoattack} {
  short=AA,
  long=AutoAttack,
}

\DeclareAcronym{pgd} {
  short=PGD,
  long=Projected Gradient Descent,
}

\DeclareAcronym{df} {
  short=DF,
  long=DeepFool,
}

\DeclareAcronym{cw} {
  short=C\&W,
  long=Carlini\&Wagner,
}

\DeclareAcronym{lr} {
  short=LR,
  long=logistic regression,
}

\DeclareAcronym{rf} {
  short=RF,
  long=random forest,
}

\DeclareAcronym{ldm}{
  short=LDM,
  long=latent diffusion model,
}

\DeclareAcronym{sd}{
  short=SD,
  long=stable diffusion,
}

\DeclareAcronym{sdv21}{
  short=SD-v2.1,
  long=stable diffusion-v2.1,
}

\DeclareAcronym{jtla} {
  short=JTLA,
  long=joint statistical testing across DNN layers for anomalies,
}

\DeclareAcronym{auc} {
  short=AUC,
  long=area-under-curve,
}

\DeclareAcronym{ib} {
  short=IB,
  long=information bottleneck,
}

\newcommand{\apgdce}{APGD-CE}
\newcommand{\apgdt}{APGD-t}

\newcommand{\squaredef}{Squares}



%% file: content/abstract.tex
In recent years, diffusion models (DMs) have drawn significant attention for their success in approximating data distributions, yielding state-of-the-art generative results.
Nevertheless, the versatility of these models extends beyond their generative capabilities to encompass various vision applications, such as image inpainting, segmentation,  adversarial robustness, among others.
This study is dedicated to the investigation of adversarial attacks through the lens of diffusion models. 
However, our objective does not involve enhancing the adversarial robustness of image classifiers.
Instead, our focus lies in utilizing the diffusion model to detect and analyze the anomalies introduced by these attacks on images.
To that end, we systematically examine the alignment of the distributions of adversarial examples when subjected to the process of transformation using diffusion models.
The efficacy of this approach is assessed across CIFAR-10 and ImageNet datasets, including varying image sizes in the latter.
The results demonstrate a notable capacity to discriminate effectively between benign and attacked images, providing compelling evidence that adversarial instances do not align with the learned manifold of the DMs.


\noindent {\small \textit{Keywords: Adversarial examples, diffusion models}}

%% file: content/content.tex
\section{INTRODUCTION}
A longstanding problem of deep learning (DL) is the vulnerability to adversarial examples \cite{szegedy2013intriguing,goodfellow2014explaining}.
These instances are maliciously crafted by introducing imperceptible perturbations to natural examples, inducing in this way erroneous predictions in DL models, such as misclassifications.

Given the potential security threats posed by the lack of adversarial robustness in real-world applications, substantial efforts have been dedicated to developing defenses against adversarial examples.
Various strategies have been explored, including addressing obfuscated gradients \cite{athalye2018obfuscated}, adversarial training  (AT) \cite{madry2017towards,zhang2019theoretically,tang2021robustart}, image denoising \cite{song2017pixeldefend,samangouei2018protecting}, and certified defences \cite{raghunathan2018certified,wong2018provable,cohen2019certified}.

The exploration of adversarial attacks on larger image dimensions is progressing rapidly, as evidenced by recent studies \cite{tang2021robustart,zhang2022beyond,chen2023diffusion}.
This progress is particularly focused on enhancing transferability to broaden their scope of application.
In contrast, defenses \cite{athalye2018synthesizing,tramer2020adaptive} targeting adversarial examples \cite{dong2020benchmarking,croce2022evaluating} concentrate on adjusting input or hyperparameters during test-time. 
However, these methods are restricted by computational demands, limiting their effectiveness to smaller image sizes (resolutions), such as CIFAR-10 \cite{krizhevsky2009learning}.
\begin{figure}[tb]
    \centering
    \includegraphics[width=\linewidth]{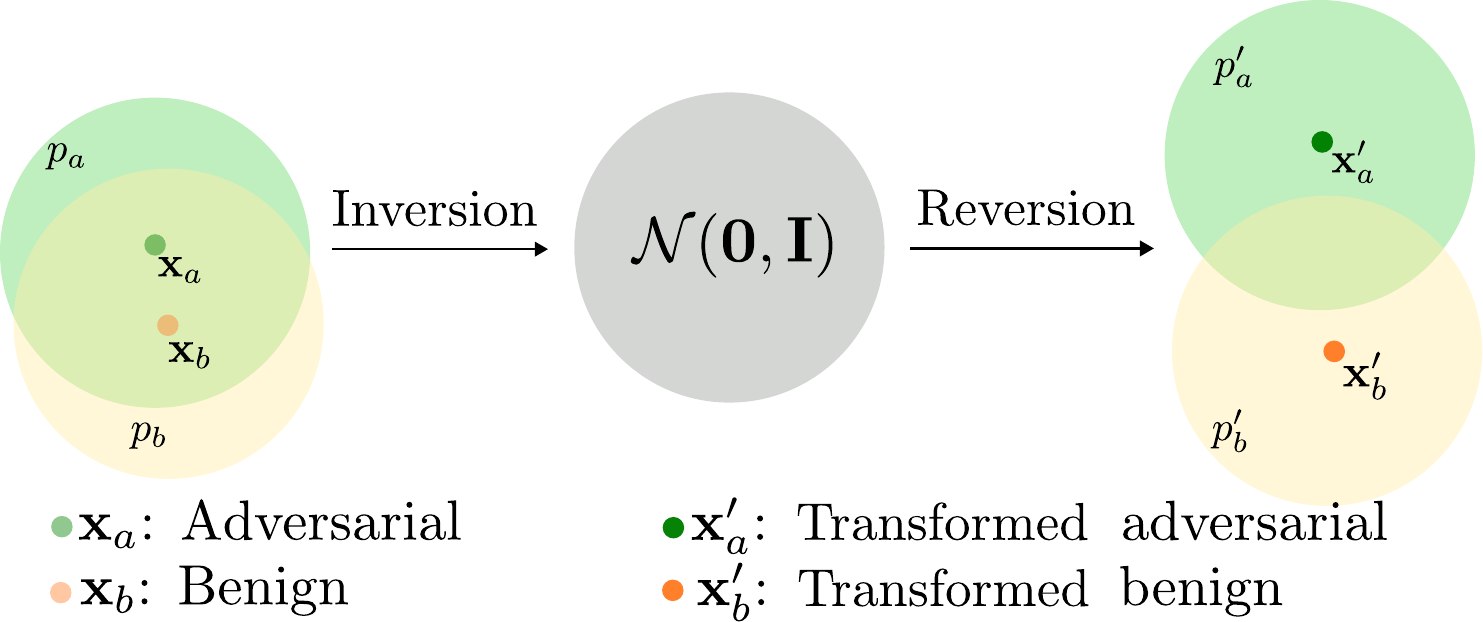}
    \caption{
        Illustration of the difference between an adversarial and a benign sample when subjected to the transformation process. 
        $p_a(\vx)$ represents the distribution of adversarial images, while $p_b(\vx)$ represents the distribution of benign images.
        Using the inversion and reversion process of a pre-trained DDIM \cite{song2020denoising}, $\vx_a$ and $\vx_b$ become $\vx_a'$ and $\vx_b'$, respectively.
        These transformed counterparts now belong to distinct distributions, namely $p_a'(\vx)$ and $p_b'(\vx)$, characterized by a significantly reduced overlap.
        Therefore, this results in a distinct representation of adversarial samples compared to benign samples.
    }
    \label{fig:teaser}
\end{figure}
Common defense methods are certifiable robustness \cite{xiao2022densepure}, randomized smoothing \cite{zhang2023diffsmooth}, and adversarial training \cite{dong2020benchmarking}.
Consequently,  supervised learning defense methods are more successful on larger image sizes but only show a proof-of-concept of the detection capabilities rather than being a defense.

Diffusion models (DMs) have emerged as a powerful family of generative models, with Denoising Diffusion Probabilistic Models (DDPMs) \cite{sohl2015deep, ho2020denoising} standing as pioneers in this field.
DDPMs have set up a new paradigm in image generation, showcasing a robust capability to produce high-quality images \cite{ricker2022towards}.
Another important model is the Denoising Diffusion Implicit Model (DDIM) \cite{song2020denoising}, known for speeding up the generation and improving sample quality.
Recently, DMs have found applications within the adversarial attack domain.
In particular, Nie et al. \cite{nie2022diffusion} developed a DM-based method to purify the input images, i.e., removal of noise or adversarial perturbations, from adversarial examples before entering a classifier. 
Expanding upon this, a recent study by Chen et al. \cite{chen2023robust} introduces a generative classifier constructed from a pre-trained DM, achieving a high level of robust accuracy against norm-bounded adversarial perturbations.

However, the aforementioned approaches concentrate on perfectly purified images.
In this paper, we investigate the impact of adversarial examples after transforming the inverse and reverse processes.
As opposed to purification methods, the transformation does not have to be perfect. 
The hypothesis behind this is that transformed adversarial images yield a certain pattern (fingerprint) resulting from a shift in the manifold learned by the DM on the benign distribution.
A similar effect has been observed in Generative Adversarial Networks (GANs) \cite{santhanam2018defending}.
Our objective is to undertake a similar investigation for DMs, requiring the data to transform a pre-trained DM (see \cref{fig:teaser}).
The input image, represented by $\mathbf{x}$, undergoes an inversion process, mapping it to the noise vector $\mathbf{x}_T$ in the noise space $\mathcal{N}(0, \mathbf{I})$.
This mapping is utilized as an initialization for the subsequent process known as reverse, wherein the denoising of the image occurs from the latent space back to its original image domain.
This transformation offers a reliable pipeline for differentiating between attacked and benign images.
By training a simple binary off-the-shelf classifier on the transformed samples, it becomes possible to detect adversarial examples with ease.
To evaluate the effectiveness of the detector, we take the ImageNet \cite{deng2009imagenet} and various \ac{wb} and \ac{bb} attacks, each characterized by different hyperparameters. \\
Our main contributions can be summarized as follows:
\begin{itemize}[nosep]
    \item We utilize the diffusion model transformation process applied to both adversarial and benign samples, enabling the discrimination between attacked and benign samples. 
    This investigation includes a broad spectrum of image sizes.
    \item The employed classifier demonstrates the ability to effectively distinguish between multiple types of attacks. 
    This implies a discerning capacity to identify not only whether an image has been subjected to an attack, but also the specific nature of the attack itself.
    \item We explore the transferability of the detector by assessing its performance on other transformed images. 
\end{itemize}

\section{RELATED WORK} \label{sec:relatedwork}

In this section, we introduce adversarial attacks and defenses, and finally, we outline the contribution of DMs to adversarial robustness.

\subsection{Adversarial Attacks} \label{sec:adversarial_attacks} 
Convolutional neural networks are known to be susceptible to adversarial attacks, i.e. small perturbations of the input images that are optimized to fool the network's decision.
The most relevant ones are presented below.\\

\noindent\textbf{\acf{fgsm}\,} \cite{goodfellow2014explaining} uses the gradients of a given model to create adversarial examples.
In other words, it embodies a white-box attack, requiring full access to the model's architecture and weights.
The process involves maximizing the model's loss with respect to the input image through gradient ascent, resulting in the creation of an adversarial image $\vx_{adv}$:
\begin{equation}
{\vx}_{adv} = \vx + \epsilon \cdot \text{sign}(\nabla_{\vx} \mJ(\vx, y)),
\end{equation}
where $\vx$ is the original input, $\epsilon$ is the perturbation magnitude, $\nabla_{\vx} \mJ(\vx, y)$ is the gradient of the loss $\mJ$ with respect to the input $\vx$, and $y$ is the true label.\\

\noindent\textbf{\acf{pgd}\,} \label{sec:pgd} \cite{madry2017towards}  is a prominent attack method. 
In contrast to FGSM, PGD is iterative and adds random initialization of the perturbations in each iteration. 
The optimized perturbations are then projected onto the $\epsilon$ ball to maintain similarity between the original and attacked images in terms of $L^2$ or $L^{\infty}$ norm. 
\begin{equation}
\scriptstyle
\vx_{{adv}}^{(0)} = \vx, \quad {\vx}_{{adv}}^{(t+1)} = \text{Clip}_{\vx}(\vx_{{adv}}^{(t)} + \alpha \cdot \text{sign}(\nabla_{\vx} \mJ(\vx_{{adv}}^{(t)}, y)), \epsilon),
\end{equation}
where $t$ is the iteration index, $\alpha$ is the step size, $\text{Clip}_\vx(\cdot, \epsilon)$ ensures that the perturbation stays within the $\epsilon$-ball around $\vx$, and $\nabla_{\vx} \mJ(\vx_{{adv}}^{(t)}, y)$ is the gradient of the loss with respect to the perturbed input at iteration $t$.
In addition, random initialization and restarts are adopted to further strengthen the attack. 
An enhanced version of PGD, known as AutoPGD \cite{croce2020reliable}, presents a variant of PGD with automatic step size tuning and a refined objective function. 
AutoPGD has demonstrated superior effectiveness compared to PGD under similar attack budgets. \\

\noindent\textbf{Masked PGD\,}  \cite{xu2023patchzero} is a variation of the PGD attack, in which perturbations are confined to a specific area in the image rather than impacting the entire image.
With a simple mask, a patch region can be defined to attack.
As shown in \cref{eq:maskedpgd}, only pixels inside the patch region $[x, y, h, w]$ will be modified by the PGD:
\begin{equation} \label{eq:maskedpgd}
{\scriptstyle
\vx_{adv}^{(t+1)} = \text{Clip}_{\vx} \left( \vx_{{adv}}^{(t)} + \alpha \cdot \text{sign}(\nabla_{\vx} \mJ(\vx_{{adv}}^{(t)}, y, \bm{\theta})[patch]), \epsilon \right).
}
\end{equation}
In this context, the term patch denotes the region specified as $[x: x + h, y: y + w]$ with the provided values of $[x, y, h, w]$.
Masked PGD is capable of targeting object detectors and video classifiers by extracting gradients from their respective loss functions.\\

\noindent\textbf{\acf{autoattack}\,} \cite{croce2020reliable} consists of four consecutive attacks executed in sequence.
If a sample's prediction can not be flipped by one attack, it is handed over to the next attack method, to maximize the overall attack success rate.
The first two are parameter-free variants of \ac{pgd} \cite{madry2017towards}  one using cross-entropy loss in \apgdce~ and the other difference of logits ratio loss (DLR) in \apgdt:
\begin{equation}
\small
    \text{DLR}(\vx,
    y) = \frac{z_y - \max_{\vx\neq \vy} z_i}{z_{\pi 1} - z_{\pi 3} }.
\end{equation}
where $\pi$ is the ordering of the components of $z$ in decreasing order.
The last two attacks comprise a targeted version of the FAB attack \cite{croce2020minimally}, and the \squaredef~ attack \cite{andriushchenko2020square} which is a black-box attack. \\

\noindent \textbf{\acf{df}\,} \cite{moosavi2016deepfool} is a  non-targeted attack that finds the minimal amount of perturbation required to flip the network decision by an iterative linearization approach.
It thus estimates the distance from the input sample to the model decision boundary.\\

\noindent\textbf{\ac{cw}\,} \cite{carlini2017towards} uses a  numerical optimization approach for the input variables, the objective is to invert the network's prediction with the minimum perturbation necessary. 
This optimization process is akin to employing the $\ell$-Broyden–Fletcher–Goldfarb–Shanno (BFGS) algorithm.
The results are presented with optimization considerations for distance metrics, specifically $L^2$, $L^0$, and $L^{\infty}$ distances.\\

\noindent \textbf{DiffAttack\,} \cite{chen2023diffusion} is the first adversarial attack based on DMs \cite{ho2020denoising}.
Unlike traditional adversarial attacks that directly manipulate pixel values, DiffAttack focuses on creating human-insensitive perturbations embedded with semantic clues, making them difficult to detect.
DiffAttack crafts perturbations in the latent space of DMs, whose properties achieve very high imperceptibility and transferability.
DiffAttack leverages the DDIM inversion process \cite{nichol2021improved}, where the clean image is mapped back into the diffusion latent space by reversing the sampling process.
The image in the latent space is directly perturbed.
To create a final attacked image, the latent space must be transformed back into an image.
Image editing approaches \cite{couairon2022diffedit, mokady2023null} propose the image latent can gradually shift to the target semantic space during the iterative denoising process.\\

\noindent \textbf{Natural Evolution Strategy} \textbf{(NES)\,}  \cite{ilyas2018black} is a method used in black-box adversarial attacks on machine learning models. 
It involves estimating the gradient by averaging the confidence scores of randomly sampled nearby points and then using projected gradient descent to perturb an image of the target class until it is sufficiently close to the original image.
NES can be applied to the embedding space, which accelerates the search process for adversarial examples.
This approach has been shown to efficiently generate perturbations for a target model, making it effective in compromising the integrity of machine learning models. \\

\noindent \textbf{Bandits\,} \cite{ilyas2018prior} is a technique employed for generating adversarial examples within a black-box setting, where only limited information about the target model is accessible. 
This approach harnesses bandit optimization, a form of online optimization featuring bandit feedback, to effectively generate adversarial examples with fewer queries and higher success rates compared to existing methods. 
The Bandits attack seamlessly integrates gradient priors, which are both data-dependent and time-dependent priors, to improve the efficiency and efficacy of black-box attacks. 
Through the incorporation of bandit optimization and gradient priors, this methodology seeks to optimize the generation of adversarial examples while minimizing the requisite number of queries to compromise the target model.
The Bandits attack has demonstrated promising outcomes boosting the performance of black-box adversarial attacks, underscoring its significance as an area of research in adversarial machine learning.

\subsection{Adversarial Defenses}
In recent years, a variety of strategies have emerged to defend against adversarial attacks.
Initially, defenses focused on supervised methods for detecting adversarial examples. 
Then, adversarial training gained popularity as another supervised learning approach. 
Later, more robust defenses have been developed to counter adaptive attacks \cite{tramer2020adaptive}.
More recently, there has been a shift towards exploring defenses that are adaptive at test-time, as highlighted in the study by Croce et al. \cite{croce2022evaluating}.
However, adversarial defenses are always one step behind adversarial attacks, since adversarial attacks have a strong ability to transfer effectively across different datasets and models. \\

\noindent \textbf{Adversarial Detection} presents a computationally efficient alternative to adversarial training, focusing on distinguishing adversarial examples from benign ones to mitigate misclassifications.
One notable approach is Spectral Defense \cite{harder2021spectraldefense}, which analyzes the frequency domain representation of input images and feature maps to identify adversarial attacks.
By leveraging the magnitude spectrum and phase of Fourier coefficients, this method achieves high detection rates.
Another one is multiLID \cite{lorenz2022unfolding} which is an improvement of the LID (Local Intrinsic Dimensionality) \cite{ma2018characterizing,lu2018limitation} in terms of detection rates.
CD-VAE (Class-Dependent Variational Auto-Encoder) \cite{yang2021class} offers an alternative approach by training a variational auto-encoder to extract class-dependent information from images, enhancing adversarial detection.
CD-VAE consistently outperforms traditional approaches, including Kernel Density (KD) \cite{carlini2017adversarial}, LID  \cite{ma2018characterizing}, and Mahalanobis distance (M-D) \cite{lee2018simple}, providing valuable insights into the realm of adversarial attack detection.\\

\noindent \textbf{Adversarial Training (AT)}  \cite{madry2017towards, zhang2019theoretically, rebuffi2021fixing} might be the most effective method, which trains neural networks using adversarial augmented data.
Noteworthy benchmark leaderboards, such as ARES-Bench\footnote{\href{https://ml.cs.tsinghua.edu.cn/ares-bench}{ml.cs.tsinghua.edu.cn/ares-bench}} and RobustBench\footnote{\href{https://robustbench.github.io}{robustbench.github.io}}, are actively tracking advancements in this area.
There is a need to address the trade-off between accuracy and resilience against adversarial examples \cite{lin2022secure}.
Despite their popularity, these models often demonstrate robustness primarily to specific attacks they are trained against, exhibiting limited generalization ability to unforeseen threats, as highlighted in \cite{tramer2019adversarial,laidlaw2020perceptual}. \\

\noindent \textbf{Defenses at Test-time} differ from the aforementioned static defense methods (detection and adversarial training) as their inputs and parameters adapt during inference.
To adapt the defense parameters, Croce et al. \cite{croce2022evaluating} evaluate optimization-based methods.
The study reveals considerable difficulty in defending at test-time, with observed accuracy drops of up to 0\%.
The evaluation compromises: I) Obfuscated gradients \cite{athalye2018obfuscated} where BPDA (Backward Pass Differentiable Approximation) can be used to attack non-differentiable preprocessing-based defenses.
II) Randomness: The inclusion of randomized elements, such as Expectation over Transformation (EoT) \cite{athalye2018synthesizing}, increases the cost for attackers.
 This is significant because attacks usually presume a global perspective on the input image.\\

\noindent \textbf{Transferability of Adversarial Examples} \cite{gu2023survey}
across different model architectures or training datasets is the ability that makes these attacks so effective.
The transferability property of adversarial examples makes black-box attacks a powerful methodology, even in cases where the attacker has limited knowledge of the victim network.
Furthermore, it is an area of active research that continuously seeks to improve transferability via new methodologies such as mitigating attention shift \cite{dong2018evading}, translation invariant attacks \cite{dong2019evading}, tune variance \cite{wang2021enhancing}, more fine-grained perturbations through diffusion models like DiffAttack \cite{chen2023diffusion}, and direction tuning \cite{yang2023improving}.
In contrast, defense strategies focus on adapting during test-time.
This adaptation is necessitated by the computational complexity involved in dealing with smaller images.
In comparison, there are only a few defenses to aim to mitigate transferability, and if they focus on scaled datasets, i.e. \cite{thomas2022dynamic} or examined in \cite{croce2022evaluating}.
At this end, attackers have an easier role because they only have to lead to misclassification to be successful, whereas defenders also need to keep up the correct prediction.

\subsection{Diffusion Models for Adversarial Robustness}
DMs have been applied within the domain of adversarial robustness, demonstrating their efficacy and versatility in addressing challenges related to the security and resilience of systems against adversarial attacks.
The DiffPure approach \cite{nie2022diffusion} utilizes DMs to purify adversarial perturbations.
This purification process involves the addition of Gaussian noises to input images, followed by the denoising of the images.
Recently, Yang et al.\cite{yang2023diffusion} claim that Diffpure is still not that protective against unseen threats. 
One potential explanation for this issue is the continued emphasis on discriminative classifiers, which may not capture the underlying structure of the data distribution.
DMs have more accurate score estimation in the whole data space, where they explore a DM itself as a robust classifier.
Moreover, DMs can also contribute to improving the certified robustness in conjunction with randomized smoothing \cite{xiao2022densepure}.
Besides, the utilization of data generated by DMs has demonstrated an improvement in the performance of adversarial training \cite{rebuffi2021fixing,wang2023better}.

\section{METHOD} \label{method}

In this section, we begin with reviewing DDPMs, and the inverse and reverse process of the DDIM \cite{song2020denoising}.
Then, we present details of our method, based on \cite{wang2023dire} originated for deepfake detection, and how to capture the adversarial attack fingerprints.

\subsection{Prelimaries}\label{sec:prelimanries}
In the following, we use the notations from DDIM \cite{song2020denoising} because we use this architecture throughout this paper and are also able to use the pre-trained weights from the DDPM architecture \cite{ho2020denoising}: We note that in  \cite{ho2020denoising}, a diffusion hyperparameter  $\beta_t$ is first introduced, and then relevant variables $\alpha_t := 1-\beta_t$ and $\bar{\alpha}_t = \prod_{t=1}^{T}\alpha_t$ are defined.
From the DDIM paper we use the notation $\alpha_t$ to represent $\bar{\alpha}_t$  and also define 
\begin{equation}
	\beta_t = 1 - \frac{\alpha_t}{\alpha_{t-1}}.
\end{equation}

\noindent \textbf{Denoising Diffusion Probabilistic Model (DDPM)} is initially proposed in \cite{sohl2015deep}, inspired by non-equilibrium thermodynamics.
This innovation has marked a significant advancement in image generation, yielding noteworthy results \cite{ho2020denoising, nichol2021improved, dhariwal2021diffusion, rombach2022high}.
DDPMs define a Markov chain of diffusion steps, progressively introducing Gaussian noise to the data.
This iterative process continues until the data transforms, ultimately converging into an isotropic Gaussian distribution. 
This defines the forward process of DMs as:
\begin{equation}
    \small
    q(\vx_t | \vx_{t-1})  = \mathcal{N} \left (\vx_t; \sqrt{\frac{\alpha_t}{\alpha_{t-1}}} \vx_{t-1}, (1 - \frac{\alpha_t}{\alpha_{t-1}})\mathbf{I} \right ),
\end{equation}
in which $\vx_t$ denotes the noisy image at the $t$-th step and let $\alpha_1, \dots, \alpha_T \in (0,1]^T$ be a predetermined decreasing schedule,  where $T$ represents the total number of steps.
An essential property conferred by the Markov chain is the direct derivation of $x_t$ from $x_0$, as follows:
\begin{equation}
\small
    q(\vx_t | \vx_{0})  = \mathcal{N} \left (\vx_t; \sqrt{\alpha_t} \vx_0, (1-\alpha_1)\mathbf{I} \right),
\end{equation}
Next, the models are trained to reverse this diffusion process, enabling the generation of samples from the noise—a process termed the {reverse process}.
The reverse process in \cite{ho2020denoising} is also defined as a Markov chain:
\begin{equation}
\small
    p_\theta(\vx_{t-1} | \vx_{t})  = \mathcal{N}(\vx_{t-1}; \boldsymbol{\mu}_{{\theta}}(\vx_t, t), \boldsymbol{\Sigma}_\theta(\vx,t)).
\end{equation}
DMs employ  a network $p_{\bm{\theta}}(\vx_{t-1} | \vx_t)$ to approximate the real distribution  $q(\vx_{t-1} | \vx_t)$. 
The primary goal of optimization is to achieve a sampling and denoising process as outlined below:
\begin{equation}
\small
L_{\text{simple}}(\bm{\theta}) = \mathbb{E}_{t, \vx_0, \bm{\epsilon}} \left [ \left \| \bm{\epsilon} - \bm{\epsilon}_{\bm{\theta}} (\sqrt{\alpha_t} \vx_0 + \sqrt{1-\alpha_t} \bm{\epsilon},t) \right \|^2 \right ].
\end{equation}
where $\bm{\epsilon} \sim \mathcal{N}(0,\mathbf{I})$. \\

\begin{figure}[t]
    \centering\includegraphics[width=\columnwidth]{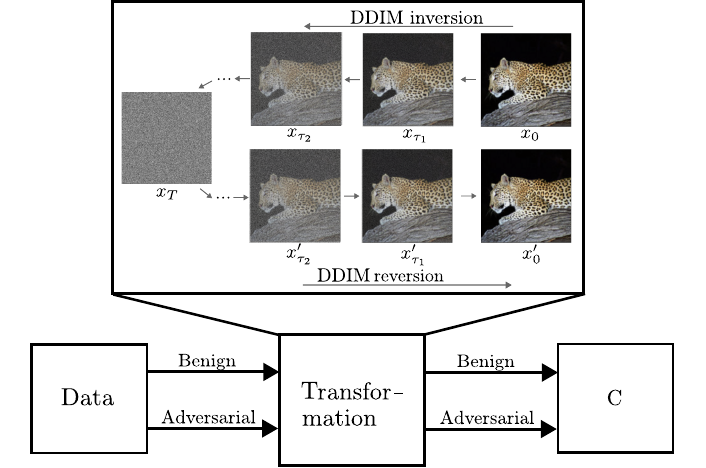}
    \caption{
        Illustration from the data generation over the transformation through a pre-trained DDIM to train a binary classifier $C$.
        Adversarial and benign samples are separately transformed.
        The transformation implies that the input image $\mathbf{x}_0$ is first gradually inverted into a noise image $\mathbf{x}_T$ using DDIM inversion  \cite{song2020denoising}, and then it is denoised step by step until the transformed $\mathbf{x}_0'$ is obtained, as illustrated in \cref{eq:reversed}.
    }
    \label{fig:ddimrecons}
\end{figure}

\noindent\textbf{Denoising Diffusion Implicit Model (DDIM)} \cite{song2020denoising} proposes a method for accelerating the iterative process without the Markov hypothesis. 
The modified reverse process in DDIM is defined as follows:
\begin{equation} \label{eq:eulerintegration}
\begin{split}
\vx_{t-1} &= \sqrt{\alpha_{t-1}} \left ( \frac{\vx_t - \sqrt{1-\alpha_t} \bm{\epsilon}_{\bm{\theta}} (\vx_t, t)}{\sqrt{\alpha_t}} \right ) + \\
& \sqrt{1-\alpha_{t-1} - \sigma^2_t} \cdot \bm{\epsilon}_{\bm{\theta}} (\vx_t, t) + \sigma_t \bm{\epsilon}_t.
\end{split}
\end{equation}
In the given equation, when $\sigma_t = 0$, the term involving $\sigma_t \bm{\epsilon}_t$ becomes zero. 
In this case, the reverse process becomes deterministic (backward process) because the term involving noise, $\sigma_t$ $\bm{\epsilon}_t$, is zero. 
In a deterministic process, each input uniquely determines the corresponding output. 
Therefore, when $\sigma_t = 0$, the reverse process is fully determined by the given formula, and there is no randomness introduced during the backward process.
Furthermore, when $T$ is large enough (e.g. $T=1000$), \cref{eq:eulerintegration} can be seen as Euler integration for solving Ordinary Differential Equations (ODEs):
\begin{equation}
\small
\frac{\vx_{t - \Delta t}}{\sqrt{\alpha_t - \Delta t}} = \frac{\vx_t}{\sqrt{\alpha_t}} + \left( \sqrt{\frac{1-\alpha_{t - \Delta t}}{\alpha_{t - \Delta t}}} - \sqrt{\frac{1-\alpha_t}{\alpha_t}} \right) \bm{\epsilon}_{\bm{\theta}}(\vx_t, t).
\end{equation}
\noindent Suppose $\sigma = \sqrt{1-\alpha}/\sqrt{\alpha}, \bar{\vx} = \vx / \sqrt{\alpha}$, the corresponding ODE becomes:
\begin{equation}
\small
d\bar{\vx}(t) = \bm{\epsilon}_{\bm{\theta}} \left( \frac{\bar{\vx}(t)}{\sqrt{\sigma^2+1}}, t \right) d\sigma(t).
\end{equation}
Then, the inversion process (from $\vx_t \text{ to } \vx_{t+1}$) is then later reversed:
\begin{equation}
\small
\frac{\vx_{t+1}}{\sqrt{\alpha_t + 1}} = \frac{\vx_t}{\sqrt{\alpha_t}} + \left( \sqrt{\frac{1-\alpha_{t+1}}{\alpha_{t+1}}} - \sqrt{\frac{1-\alpha_t}{\alpha_t}} \right) \bm{\epsilon}_{\bm{\theta}}(\vx_t, t). \label{eq:reverse}
\end{equation}
This procedure aims to acquire the corresponding noisy sample $\vx_T$ for an input image $\vx_0$. 
Nevertheless, performing step-by-step inversion or sampling is notably time-consuming.
To speed up the DM sampling, 
DDIM \cite{song2020denoising} permits us to sample a subset of $S$ steps $\tau_1, \dots, \tau_S$, so that the neighboring $\vx_t$ and $\vx_{t+1}$ become $\vx_{\tau_t}$ and $\vx_{\tau + 1}$, respectively, in  \cref{eq:eulerintegration} and  \cref{eq:reverse}.

\subsection{Method Details}
Given an input image $\vx_0$, our objective is to classify whether it is adversarial or natural (benign).
To achieve this, we utilize a pre-trained diffusion model, specifically a DDIM, and apply the inversion process, gradually introducing Gaussian noise (refer to \cref{eq:reverse}).
After $T$ steps, $\vx_0$ transforms into $\vx_T$, which now belongs to an isotropic Gaussian noise distribution. 
Subsequently, we apply the reverse process (refer to \ref{eq:eulerintegration}) to convert the noisy image, resulting in a recovered version $\vx$.
The overall transformation is defined as:
\begin{equation}
   \vx_0' = \mathbf{R}(\mathbf{I}(\vx_0)), \label{eq:reversed}
\end{equation}
where $\mathbf{I}(\cdot)$ represents the inversion process, and $\mathbf{R}(\cdot)$ denotes the reverse process. 

To train a binary classifier differentiating between adversarial and benign samples, we apply this transformation to both types of samples.
The outcomes are then used to train the binary classifier using binary cross-entropy loss, formulated as:
\begin{equation}
    \ell(y, y') = - \sum_{i=1}^{N} \left (  y_i \log (y'_i) + (1-y_i) \log(y'_i)  \right ), 
\end{equation}
where $N$ is mini-batch size, $y$ is the ground-truth label, and $y'$ is the corresponding prediction by the detector. 

\section{EXPERIMENTS}
In this section, we begin by introducing the datasets, metrics, and the training procedure. 
Following this, we present and discuss an extensive collection of experiments.

\subsection{Datasets} \label{sec:dataset}
For our experiments, we create several datasets: Adversarial datasets corresponding to an attack denoted as $\mathcal{X}_{\text{a}}$ and a benign dataset denoted as $\mathcal{X}_{\text{b}}$.
The adversarial datasets are crafted with a batch size of 50, giving particular attention to black-box attacks due to their enhanced performance with larger batch sizes \cite{sun2022towards}.
Specifically, we only take images where the applied attacks are successful.
To create the datasets, we systematically gather 10,000 benign datasets and 10,000 datasets subjected to adversarial attacks, ensuring a complete absence of overlap between the two.
Ultimately, we partition the datasets into training 80\%, validation 10\%, and test 10\% sets.\\

\noindent\textbf{CIFAR-10} \cite{krizhevsky2009learning}:
We employ the CIFAR-10 dataset as our low-resolution dataset (size $32\times 32$ pixels). 
The reverse process is performed using the DDIM CIFAR-10 L-hybrid model, accessible at the DDIM repository\footnote{\href{https://github.com/openai/improved-diffusion}{github.com/openai/improved-diffusion}, CIFAR-10 L-hybrid}.\\

\noindent\textbf{ImageNet} \cite{deng2009imagenet}:
We utilize the ImageNet dataset as our foundational dataset.
To ensure a class-balanced representation, we curate a dataset by extracting 100 samples from each of the 100 classes, resulting in a total of 10,000 samples.
The image sizes are chosen to align with the pre-trained unconditional diffusion models, specifically for sizes $256\times 256$\footnote{\href{https://github.com/openai/guided-diffusion}{github.com/openai/guided-diffusion}}, and $512\times 512$ pixels\footnote{\href{https://huggingface.co/lowlevelware/512x512_diffusion_unconditional_ImageNet}{huggingface.co/lowlevelware/512x512\_diffusion\_unconditional}}.\\

\noindent\textbf{Compressed ImageNet} \cite{pytorch-playground}:
In the context of the DiffAttack \cite{chen2023diffusion}, we leverage the Compressed ImageNet dataset \cite{zhang2022beyond}, as this attack has been optimized and previously assessed in other studies for adversarial robustness. 
The attack involves downsampling the input image from $256\times 256$ to $224\times 224$ pixels.
To meet the required image size of $256\times 256$ pixels for DM reverse steps, we employ zero-padding to the left and bottom of the image.

\subsection{Evaluation Metrics}
To comprehensively evaluate the efficacy of our approach, we employ a diverse set of standard metrics commonly utilized in detection scenarios. 
These metrics serve as quantitative measures, offering insights into the robustness and accuracy of the proposed method.
In particular, we use the following metrics:
area-under-curve (AUC), average precision (AP), true negative rate (TNR), false negative rate (FNR), true positive rate (TPR), false positive rate (FPR), precision, recall, and F1.

\begin{algorithm}[tb]
    \caption{Training of the adversarial detector.} \label{alg:training}
    \begin{algorithmic}[1]
    \Require Benign dataset $\mathcal{X}_{\text{b}}$, Adversarial dataset $\mathcal{X}_{\text{a}}$
    \Ensure  Trained classifier $C$ 
    \State Generate transformed dataset $\mathcal{X}_{\text{tf}}$ using a pre-trained diffusion model
    \State Split $\mathcal{X}_{\text{tf}}$ into training set $\mathcal{X}_{\text{train}}$ and test set $\mathcal{X}_{\text{test}}$
    \State Initialize ResNet-50 model $C$
    \State Train $C$ using $\mathcal{X}_{\text{train}}$
    \State Evaluate $C$ using $\mathcal{X}_{\text{test}}$
    \State\Return $C$
    \end{algorithmic}
\end{algorithm}

\subsection{Training Procedure}
In \cref{alg:training}, we outline the training procedure for the adversarial detector. 
The initial step involves generating both adversarial and benign datasets (see in \cref{sec:dataset}). 
Then, we apply the transformation, using DM, to all data samples.
Note that this procedure necessitates the use of specific dimensions, such as $32\times 32$, $256\times 256$, and $512\times 512$ pixels, as the pre-trained DM is designed to process images of these sizes.
The resulting transformed images are used for training the classifier $C$, employing either ResNet-50, originally pre-trained on ImageNet, or ResNet18\footnote{\href{https://huggingface.co/edadaltocg/resnet18\_cifar10}{huggingface.co/edadaltocg/resnet18\_cifar10}}, originally pre-trained on CIFAR-10.
Throughout the training of the classifier, we maintain the respective image sizes, except for the transformed images of $256\times 256$ pixels, which are randomly cropped to the dimensions of $224\times 224$ to align with previous work.
We pre-process the data accordingly:
I) During training, the images fed into the network are randomly cropped and horizontally flipped with a probability of $0.5$.
II) During testing, the images are center-cropped.

\subsection{Results and Discussion}
This subsection offers a comprehensive analysis of the results from the proposed approach.
We delve into a detailed discussion of the performance evaluation, with a specific focus on addressing a fundamental question: 
\textit{Can the proposed methodology effectively distinguish between instances classified as under attack and those labeled as benign across various image resolutions?}

\begin{table}[tb]
\caption{
CIFAR-10. The table displays the classification accuracy of our method along with a comparison to various types of adversarial defenses, including Adversarial Training (AT), Input Purification (IP), Supervised Learning (SL), and Auxiliary Network (AN).
Attack methods are AA and PGD with $\epsilon=4/255$; and AA$^*$ and PGD$^*$ with $\epsilon=8/255$.
} \label{tab:resultscifar10}
\resizebox{\columnwidth}{!}{
    \begin{tabular}{ll|llllll}
    \toprule[1pt]
    \hline
    \textbf{Method} & \textbf{Type} & \textbf{Attack} & \textbf{ACC} & \textbf{AUC} & \textbf{TNR} & \textbf{$\mathbb{\mathcal{A}}_\text{std}$} & \textbf{$\mathbb{\mathcal{A}}_\text{rob}$} \\ \midrule
    AT-DDPM-$\ell_\infty$ \cite{rebuffi2021fixing} & AT     & AA  & 76.08 & - & -     & 88.87 & 63.28 \\
    AT-EDM-$\ell_\infty$ \cite{wang2023better}     & AT     &     & 82.13 & - & -     & 93.36 & 70.91 \\
    RaWideResNet-70-16 \cite{peng2023robust}       & AT     &     & 82.17 & - & -     & 93.27 & 71.07 \\
    Visual Prompting \cite{chen2023visual}         & AT     &     & 46.16 & - & -     & 57.57 & 34.75 \\
    RDC \cite{chen2023robust}                      & IP     &     & 83.23 & - & - & 93.16 & 73.24 \\
    DiffPure  \cite{nie2022diffusion}              &        &     &       &   &   &       &  \\
   \hspace{0.3cm} WRN-28-10                        & IP     &     & 79.83 & - & - & 89.02 & 70.64 \\
   \hspace{0.3cm} WRN-70-16                        & IP     &     & 80.68 & - & - & 90.07 & 71.29 \\
    HEDGE \cite{wu2021attacking}                   & IP     & AA  & 79.62 & - & - & 89.16 & 70.07 \\
                                                   & IP     & PGD & 79.11 & - & - & 89.16 & 69.04 \\
    AID Purifier \cite{hwang2023aid}               & IP, AN &     & 70.42 & - & - & 88.28 & 52.56 \\
    Mao et.al.\cite{mao2021adversarial}            &        &     & 64.21 & - & - & 60.67 & 67.79 \\
    multiLID \cite{lorenz2022unfolding}            & SL, AN & AA  & 96.43 & 99.37 & 94.11 & 99.86 & \textbf{99.93} \\
                                                   & SL, AN & PGD & 93.93 & 97.94 & 92.86 & 92.86 & 95.22  \\
    SD$_\text{WB}$ \cite{lorenz2021detecting}      & SL, AN &    & 97.25 & \textbf{99.93} & 95.14 & 99.54 & 95.08  \\
    SD$_\text{BB}$ \cite{lorenz2021detecting}      & SL, AN &    & 95.53 &         99.74  & 90.99 & \textbf{100} & 91.01 \\
    CD-VAE \cite{yang2021class}                    &   &   &   &   &   &   &  \\
    \hspace{0.3cm} KD (R(x))                       & SL, AN  &    &   -   & 99.30 & 96.56 & - & - \\
    \hspace{0.3cm} LID (R(x))                      & SL, AN  &    &   -   & 97.57 & 87.54 & - & - \\
    \hspace{0.3cm} M-D (R(x))                      & SL, AN  &    &   -   & 99.79 & \textbf{99.13} & - & - \\ \midrule
    \rowcolor[HTML]{EFEFEF} 
    Ours                                           & SL, AN & AA  & \textbf{97.40} & 99.73 & 97.51 & 97.54 & 97.32 \\
    \rowcolor[HTML]{EFEFEF} 
    Ours                                          & SL, AN & PGD & 95.40 & 99.03 & 96.11 & 96.08 & 94.81 \\ \hline \bottomrule[1pt]
    \end{tabular}
    }
\end{table}

\begin{table}[t]
    \caption{
        ImageNet with 224 pixels. 
        The table displays the classification accuracy of our method along with a comparison to various types of adversarial defenses, including Adversarial Training (AT), Input Purification (IP), Supervised Learning (SL), and Auxiliary Network (AN).
        Attack methods are AA and PGD with $\epsilon=4/255$; and AA$^*$ and PGD$^*$ with $\epsilon=8/255$.
        We refer to other epsilon sizes of our proposed method on the other \cref{tab:binaryclassification}.
    } \label{tab:results}
\resizebox{\columnwidth}{!}{
\begin{tabular}{ll|llllll} 
    \toprule[1pt]
    \hline
    \textbf{Method} & \textbf{Type} & \textbf{Attack} & \textbf{ACC} & \textbf{AUC} & \textbf{TNR} & \textbf{$\mathbb{\mathcal{A}}_\text{std}$} & \textbf{$\mathbb{\mathcal{A}}_\text{rob}$} \\ 
    \hline
    Swin-L \cite{liu2023comprehensive}       & AT & AA  & 69.24 & - & - & 78.92 & 59.56 \\
    Conv-Next-L \cite{liu2023comprehensive}  & AT &     & 68.25 & - & - & 78.02 & 58.48 \\
    DiffPure \cite{nie2022diffusion}         &    &     &  &  &  &  &  \\
    \hspace{0.3cm} ResNet-50                 & IP & PGD & 54.36 & - & - & 67.79 & 40.93 \\
    \hspace{0.3cm} WRN-50-2                  & IP &     & 57.78 & - & - & 71.16 & 44.39 \\
    \hspace{0.3cm} DeiT-S                    & IP &     & 58.41 & - & - & 73.63 & 43.18 \\
    multiLID \cite{lorenz2022unfolding}      & SL, AN & AA*   & 99.46 & 99.98 & 99.29 & 98.91 & 99.64 \\
    multiLID \cite{lorenz2022unfolding}      & SL, AN & PGD*  & 89.29 & 95.45 & 89.46 & 97.93 & 89.11 \\
    $\text{SD}_\text{WB}$ \cite{lorenz2021detecting}  & SL, AN & AA & 97.12 & 99.71 & 96.75 & 97.51 & 96.75 \\
    $\text{SD}_\text{BB}$ \cite{lorenz2021detecting}  & SL, AN  &    & 83.27 & 83.27 & 59.25 & 91.04 & 59.25 \\
    CD-VAE \cite{yang2021class} &   &  &  &  &  &  &  \\
    \hspace{0.3cm} KD  ($R(\vx)$)  & SL, AN & PGD & - & \textbf{100} & 96.56 & - & - \\
    \hspace{0.3cm} LID ($R(\vx)$)  & SL, AN &     & - & 97.38 & 87.54 & - & - \\
    \hspace{0.3cm} M-D ($R(\vx)$)  & SL, AN &     & - & 99.77 & 99.13 & - & - \\ 
    \midrule 
    \rowcolor[rgb]{0.937,0.937,0.937} Ours  & SL, AN & AA & 95.75 & 99.63 & 94.0 & 94.0 & 97.5 \\
    \rowcolor[rgb]{0.937,0.937,0.937} Ours  & SL, AN & PGD &\textbf{99.1} & 99.96 & \textbf{99.7} & \textbf{99.7} & \textbf{98.54} \\
    \hline \bottomrule[1pt]
\end{tabular}
}
\end{table}

In \cref{tab:resultscifar10}, we conduct a comparative analysis of our method against various adversarial defenses, encompassing supervised learning (i.e. adversarial training), and input purification as principal components. 
Our method aligns with the supervised learning model category, which usually is determined as the first proof-of-concept.
For our evaluations, we employ AA and PGD attacks, considering their prominence in related work and the ongoing debate on their efficacy in assessing the robustness of  DM-based defenses \cite{lee2023robust}.
Supervised learning (SL) methods typically serve as an initial benchmark to assess the capabilities of a method for a specific learning task, often yielding superior results compared to alternative approaches.
Consequently, different evaluation metrics are employed, leading to nuanced comparisons.
However, it is essential to note that other methods may have distinct focuses, such as evaluating robustness during test-time \cite{chen2023visual}, resilience against unseen threats \cite{laidlaw2020perceptual}, or flexible to adaptive attacks \cite{croce2022evaluating}.
Due to the divergent focuses, evaluations are often constrained to low-resolution datasets, such as CIFAR-10.

Turning our attention to a higher resolution dataset, ImageNet, in \cref{tab:results} we observe that, apart from SL methods, the defense performance decreases when concentrating on static defense mechanisms without accounting for unseen threats.
Notably, while multiLID \cite{lorenz2022unfolding} and SD \cite{lorenz2021detecting} represent straightforward defense strategies, CD-VAE \cite{yang2021class} emerges as a more intricate method by using the common information per class extracted from a GAN.

Furthermore, we provide an extensive examination of the ImageNet dataset, exploring various adversarial attacks.
In this analysis, we aim to assess the generalizability of the employed method presented in \cref{tab:binaryclassification}.
This involves assessing its performance across various attacks and varying the hyperparameter $\epsilon$ sizes when applicable.
Additionally, we consider DiffAttack, a novel distance metric-based attack meticulously optimized for the compressed ImageNet dataset and known for its invisible perturbations.
Continuing the evaluation, we extend the same rigorous procedure to black-box attacks, to gain insights into the method's resilience across various adversarial scenarios. 
Remarkably, our approach consistently demonstrates promising detection outcomes, showing its efficacy in mitigating diverse types of attacks.

This variation of methods, attacks, and datasets ensures a comprehensive evaluation of method performance across different scales and contexts.
Nevertheless, we observe that adversarial examples, characterized by subtle pixel changes, impact the DM transformation (see \cref{app:reconstruction} and \ref{app:evolution}).
This influence extends to the DM's reverse process, leading to the emergence of identifiable and learnable patterns.

\subsection{Ablation Study}
In this section, we present an ablation study to evaluate the pattern capabilities derived from the transformation (inverse and reverse steps of DDIM).
Our study focuses on investigating the following questions:
\textit{I) How many reverse steps are required for uncovering the adversarial examples?
II) Can these adversarial examples be identified as a unique fingerprint, and what are the transferability properties?
}

\begin{table*}[tb]
\caption{
The table presents the classification accuracy on the ImageNet dataset.
We use the pre-trained ResNet-18 model on ImageNet.
*Compressed Imagenet \cite{pytorch-playground}.
The attacks are generated on a batch size of 50.
} 
\label{tab:binaryclassification}
\centering
\resizebox{\textwidth}{!}{
    \begin{tabular}{llllllllllllll} 
        \toprule[1pt]
        \hline
        \textbf{Attack} & $\epsilon$ & \textbf{AUC} & \textbf{ACC} & \textbf{AP} & $\mathbb{\mathcal{A}}_\text{std}$ & $\mathbb{\mathcal{A}}_\text{rob}$ & \textbf{TNR} & \textbf{FNR} & \textbf{TPR} & \textbf{FPR} & \textbf{Prec} & \textbf{Rec} & \textbf{F1} \\ 
        \midrule
        \multicolumn{14}{c}{\textbf{\Acl{wb}}} \\ 
        \midrule
        FGSM & 1/255 & 99.3 & 96.3 & 99.48 & 95.7 & 96.9 & 95.7 & 3.1 & 96.9 & 4.3 & 95.75 & 96.9 & 96.32 \\
        FGSM & 2/255 & 99.96 & 99.15 & 99.97 & 99.0 & 99.3 & 99.0 & 0.7 & 99.3 & 1.0 & 99.0 & 99.3 & 99.15 \\
        FGSM & 4/255 & 99.88 & 98.65 & 99.87 & 98.7 & 98.6 & 98.7 & 1.4 & 98.6 & 1.3 & 98.7 & 98.6 & 98.65 \\
        FGSM & 8/255 & 100   & 99.95 & 100   & 100 & 99.9 & 100 & 0.1 & 99.9 & 0.0 & 100 & 99.9 & 99.95 \\
        PGD  & 1/255 &99.09  & 93.15 & 98.32 & 91.3 & 95.0 & 91.3 & 5.0 & 95.0 & 8.7 & 91.61 & 95.0 &93.27 \\
        PGD & 2/255 & 99.84 & 98.2 & 99.92 & 98.7 & 97.7 & 98.7 & 2.3 & 97.7 & 1.3 & 98.69 & 97.7 & 98.19 \\
        PGD & 4/255 & 99.96 & 99.1 & 99.97 & 99.7 & 98.5 & 99.7 & 1.5 & 98.5 & 0.3 & 99.7 & 98.5 & 99.09 \\
        PGD & 8/255 &  100 & 99.75  & 100 & 99.8 & 99.7 & 99.8 & 0.3 & 99.7 & 0.2 & 99.8 & 99.7 & 99.75 \\
        Masked PGD & 1 & 100 & 99.75 & 99.84 & 99.8 & 99.7 & 99.8 & 0.3 & 99.7 & 0.2 & 99.8 & 99.7 & 99.75 \\
        AA & 1/255 &99.69& 96.3 & 99.55 & 96.2 & 96.4 & 96.2 & 3.6 & 96.4 & 3.8 & 96.21 & 96.4 & 96.3 \\
        AA & 2/255 &99.95& 98.65 & 99.95 & 99.2 & 98.1 & 99.2 & 1.9 & 98.1 & 0.8 & 99.19 & 98.1 & 98.64 \\
        AA & 4/255 &99.63& 95.75 & 99.58 & 94.0 & 97.5 & 94.0 & 2.5 & 97.5 & 6.0 & 94.2 & 97.5 & 95.82 \\
        AA & 8/255 &99.93& 98.65 & 99.93 & 98.9 & 98.4 & 98.9 & 1.6 & 98.4 & 1.1 & 98.89 & 98.4 & 98.65 \\
        DF & - & 98.5& 93.3& 98.59& 93.0& 93.6& 93.0& 6.4& 93.6& 7.0& 93.04& 93.6& 93.32\\
        CW & - & 95.93& 88.85& 95.84& 87.6& 90.1& 87.6& 9.9& 90.1& 12.4& 87.9& 90.1& 88.99 \\ 
        DiffAttack* & - &100 & 99.9 & 100 & 99.8 & 100 & 99.8 & 0.0 & 100 & 0.2 & 99.8 & 100 & 99.9 \\ 
        \midrule
        \multicolumn{14}{c}{\textbf{\Acl{bb}}} \\ 
        \midrule
        Square & 2/255 & 98.5 & 93.3& 98.59& 93.0& 93.6& 93.0& 6.4& 93.6& 7.0& 93.04& 93.6& 93.32 \\
        Square & 4/255 & 98.67& 96.0& 98.92& 97.3& 94.7& 97.3& 5.3& 94.7& 2.7& 97.23& 94.7& 95.95 \\
        Square & 8/255 & 99.82& 98.4& 99.84& 99.2& 97.6& 99.2& 2.4& 97.6& 0.8& 99.19& 97.6& 98.39 \\
        Bandits & 0.05 & 100  &99.65& 100  & 99.3& 100 & 99.3& 0.0& 100 & 0.7& 99.3 & 100 & 99.65\\
        NES     & 0.05 & 99.86& 98.1& 99.85& 97.9& 98.3& 97.9& 1.7& 98.3& 2.1& 97.91& 98.3& 98.1 \\
        \hline
        \bottomrule[1pt]
    \end{tabular}
}
\end{table*}
\subsubsection{Impact of the Diffusion Reverse Steps}
In DDPM architecture, the reverse process is notorious for its computational demands.
By default, we set the reverse steps to 1000 steps.
However, in this subsection, we analyze the impact of changing the number of reverse steps on the model's performance.
Results shown in \cref{tab:diff_steps} indicate that reducing the number of steps leads to a decline in accuracy.
TNR and FPR are inversely proportional.
On the other hand, if we set $T=2000$, doubling the number of reverse steps only yields a marginal improvement in accuracy. 
This implies that the standard parameter ($T=1000$) for reverse steps is appropriate.
\begin{table}[H]
\caption{
The relationship between denoising steps and model accuracy.
$T=1000$ is the default value.
}
\centering
\resizebox{\columnwidth}{!}{
\begin{tabular}{lllllllll} 
    \toprule[1pt]
    \hline
\textbf{Attack} & \textbf{Steps} & $\epsilon$ &  \textbf{ACC} & \textbf{AP} & \textbf{TNR} & \textbf{FNR} & \textbf{TPR} & \textbf{FPR}  \\ 
\hline
    PGD   & 2000  & 1/255     &  96.1& 99.53& 95.9& 3.7& 96.3& 4.1   \\
          & 1000  & 1/255     & 93.15  & 98.32 & 91.3 &  5.0 & 95.0 & 8.7   \\
          & 750   & 1/255     & 88.75           & 96.41 & 92.2 & 14.7 & 85.3 & 7.8   \\
          & 500   & 1/255     & 80.3            & 89.88 & 82.8 & 22.2 & 77.8 & 17.2  \\
          & 250   & 1/255     & 52.95           & 47.51 & 7.7  & 1.8  & 98.2 & 92.3  \\
    \hline
\bottomrule[1pt]
\end{tabular} \label{tab:diff_steps}
}
\end{table}

\subsubsection{Identification and Transferability Capability Evaluation}
\begin{figure*}[tp]
    \centering
    \includegraphics[width=1\linewidth]{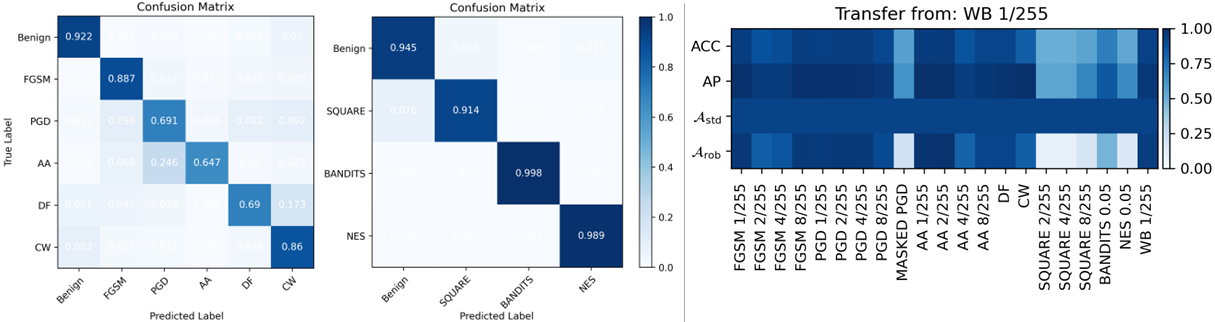}
    \caption{
    \small
        Left: Identification of with-box attacks.
        The classifier has trouble distinguishing PGD, AA, and DF.
        Benign examples can be clearly distinguished from attacked ones.
        Center: Identification of black-box attacks.
        The classifier can clearly distinguish the data transformations of each attack method.
        Right: Transferability of a binary classifier trained on white-box attacks ($\epsilon=1/255$; without Masked PGD) and tested on all other datasets, as indicated on the x-axis. 
    }
    \label{fig:transfer_wb}
\end{figure*}

In this subsection, our focus centers on investigating the transferability capabilities inherent in our method.
To address the identification question, we explore the effectiveness of our approach on the ImageNet dataset.
The \cref{fig:transfer_wb} (left) displays the confusion matrix for white-box attacks, where we can see significantly high accuracy scores for benign data. 
However, noticeable performance degradation is observed when dealing with PGD, AA, and DF attacks.
PGD and AA exhibit considerable similarity, leading to frequent confusion in their identification.
DF is commonly confused with the CW attack.
Expanding our study to black-box attacks, the confusion matrix shown in  \cref{fig:transfer_wb} (center) reveals identification results with very high accuracy scores. 
This thorough analysis not only explores how well our method works on ImageNet but also shows its strength when facing various types of attacks.

Lastly, we investigate the transferability capabilities of the binary classifier.
In general, adversarial examples have the transferability property, which makes them very powerful.
The question arises: \textit{does this property persist after the transformation?} 
We examine the transferability capabilities from one attack to another, augmenting the training dataset with diverse attacks.
In \cref{fig:transfer_wb} (right),  the detector is trained on {FGSM, PGD, AA} (1/255), DF, and CW datasets, because transferability with smaller $\epsilon$ to the same attacks but with a larger $\epsilon$ size is high.
In contrast, the transferability towards unforeseen attacks is very low.
Theoretically, the detector would rely on a huge amount of data but could be still bypassed because this detector is static.
More transferability evaluations can be found in \cref{app:transferability}.

\begin{table*}[tp]
\caption{
Evaluation of various image resolutions on the ImageNet dataset. 
}
\label{tab:scaling}
\centering
\resizebox{\textwidth}{!}{
    \begin{tabular}{lllllllllllllll} 
        \toprule[1pt]
        \hline
        \textbf{Attack} & \textbf{Size} & $\epsilon$ & \textbf{AUC} & \textbf{ACC} & \textbf{AP} & $\mathbb{\mathcal{A}}_\text{std}$ & $\mathbb{\mathcal{A}}_\text{rob}$ & \textbf{TNR} & \textbf{FNR} & \textbf{TPR} & \textbf{FPR} & \textbf{Prec} & \textbf{Rec} & \textbf{F1} \\ 
        \midrule
        PGD & $224\times 224$  & 1/255 & 99.09  & 93.15 & 98.32&91.3&95.0&91.3&5.0&95.0&8.7&91.61&95.0&93.27 \\
        & $512\times 512$  & 1/255 & 99.81 & 97.95 & 99.8  & 97.4 & 98.5 & 97.4 & 1.5 & 98.5 & 2.6 & 97.43 & 98.5 & 97.96 \\
        \hline
        \bottomrule[1pt]
    \end{tabular}
}
\end{table*}

\subsubsection{Impact of higher Image Resolutions}
In \cref{tab:scaling}, we present a comparative analysis of image resolutions, specifically 224 and 512 pixels.
To the best of our knowledge, our proposed method is pioneering, presenting the initial results in adversarial detection for an image resolution of 512 pixels.
The variability in detection accuracy across different image resolutions seems to be minimal. 
Nevertheless, it is noticeable that detection accuracy tends to increase with higher image resolution.

\section{CONCLUSION} 
In this work, we introduce an innovative approach that utilizes diffusion models to transform both adversarial and benign examples, subsequently employing these transformations to train a classifier.
Our method unveils compelling evidence suggesting that adversarial examples do not belong to the learned manifold of the diffusion model (DM), highlighting its potential to uncover adversarial examples.
We provide empirical evidence for this hypothesis and show that our proposed transformation acts as a reliable tool for uncovering adversarial perturbed samples and their fingerprints.

While our method achieves remarkable detection accuracy across diverse attacks and resolutions, it is important to acknowledge its role as a complementary defense rather than a standalone solution. 
This is due to its limitations in countering adaptive defenses capable of dynamic adjustments during test-time, as well as the constraint in the transferability to unseen threats, similar to purification approaches.

We believe that our study contributes valuable insights into the fields of adversarial robustness of DMs and also explores the learned manifolds of DMs.
The evaluation of our proposed defenses involves the utilization of attack methods based on Projected Gradient Descent or AutoAttack, among others, as well as black-box attacks on the challenging ImageNet dataset up to an image size of $512\times 512$ pixels.
Various experiments show the effectiveness and the generalization of the method across different attacks and image sizes.

%% file: content/supplementary.tex
\appendix
\onecolumn
\section*{APPENDIX}
\renewcommand{\thesubsection}{\Alph{subsection}}
\subsection{Transformation Examples} \label{app:reconstruction}
\begin{figure}[htp]
    \centering
    \includegraphics[width=1\textwidth]{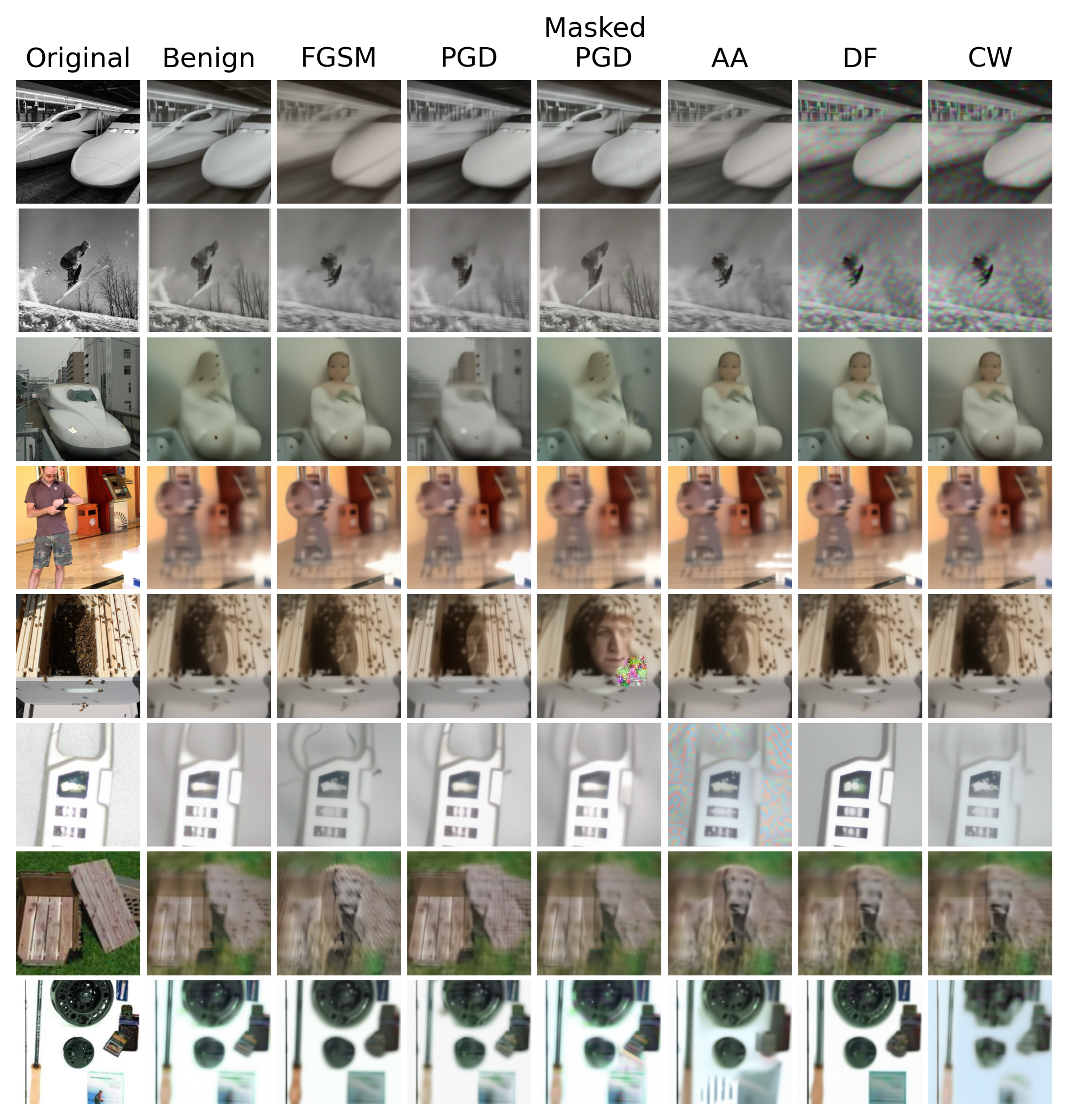}
    \caption{Transformation of white-box attacked images.}
    \label{fig:grid}
\end{figure}

\begin{figure}[htp]
    \centering
    \includegraphics[width=0.65\textwidth]{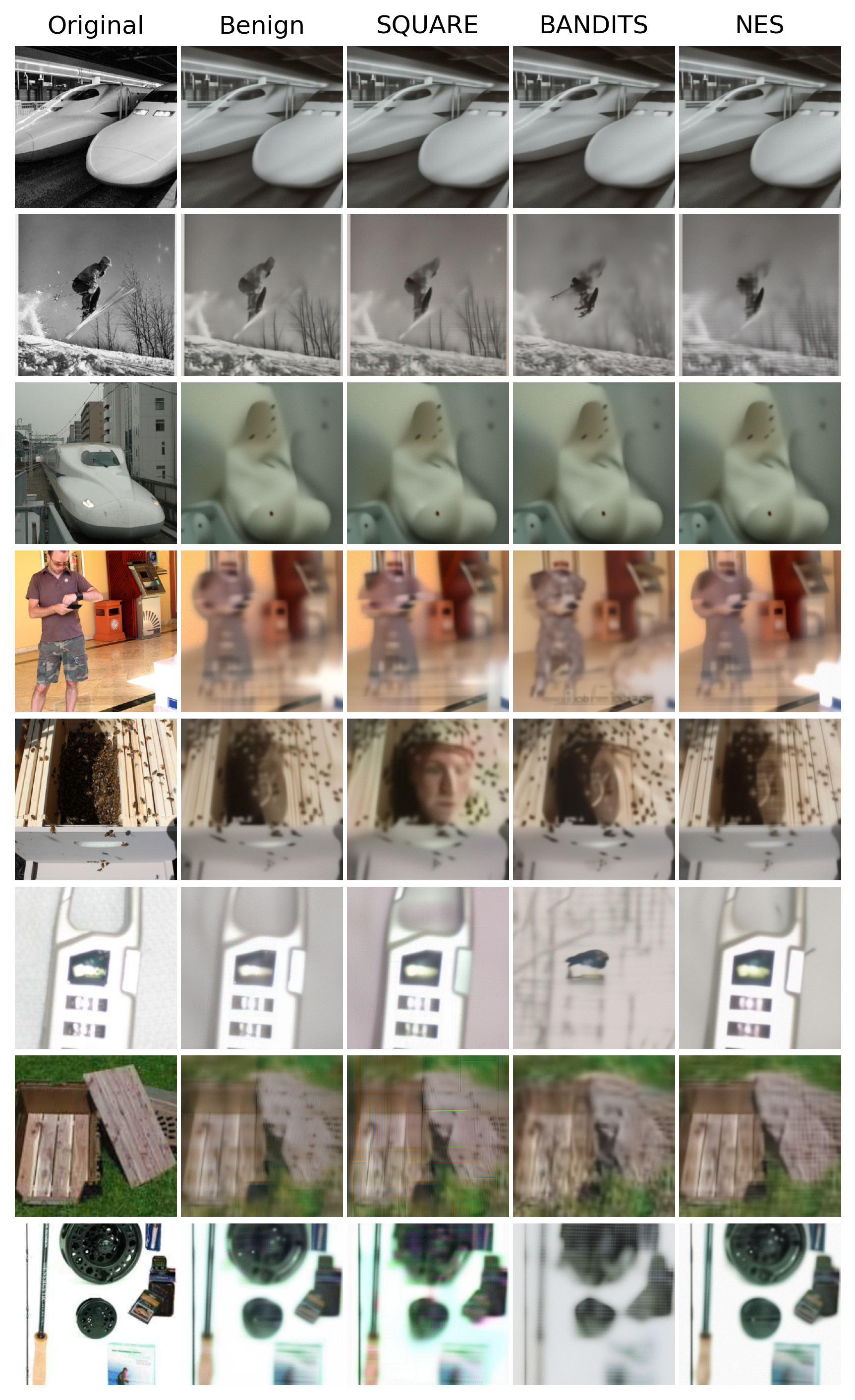}
    \caption{Transformation of black-box attacked images.}
    \label{fig:grid_bb}
\end{figure}

In this section, we depict selected transformed example images of adversarial and benign samples.
In \cref{fig:grid}  we show examples from white-box attacks and in \cref{fig:grid_bb} from black-box attacks.
Per attack, there are tiny differences after transformation or even color jitter appears.

\begin{figure}[h]
    \centering
    \includegraphics[height=\textheight]{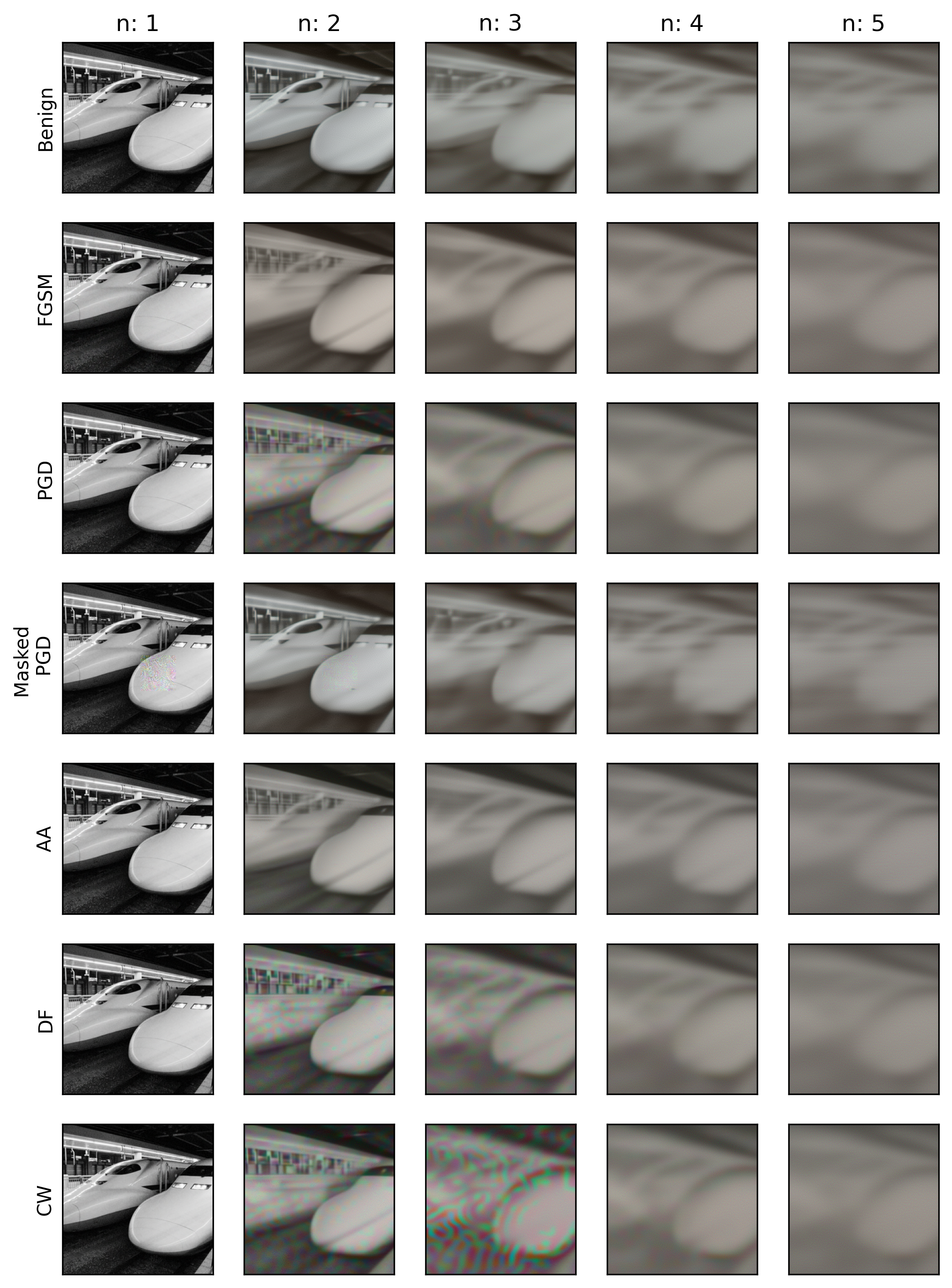}
    \caption{Recursive transformation from white-box attacked image (n: 1 to n: 5).}
    \label{fig:evolution}
\end{figure}

\begin{figure}[h]
    \centering
    \includegraphics[height=0.6\textheight]{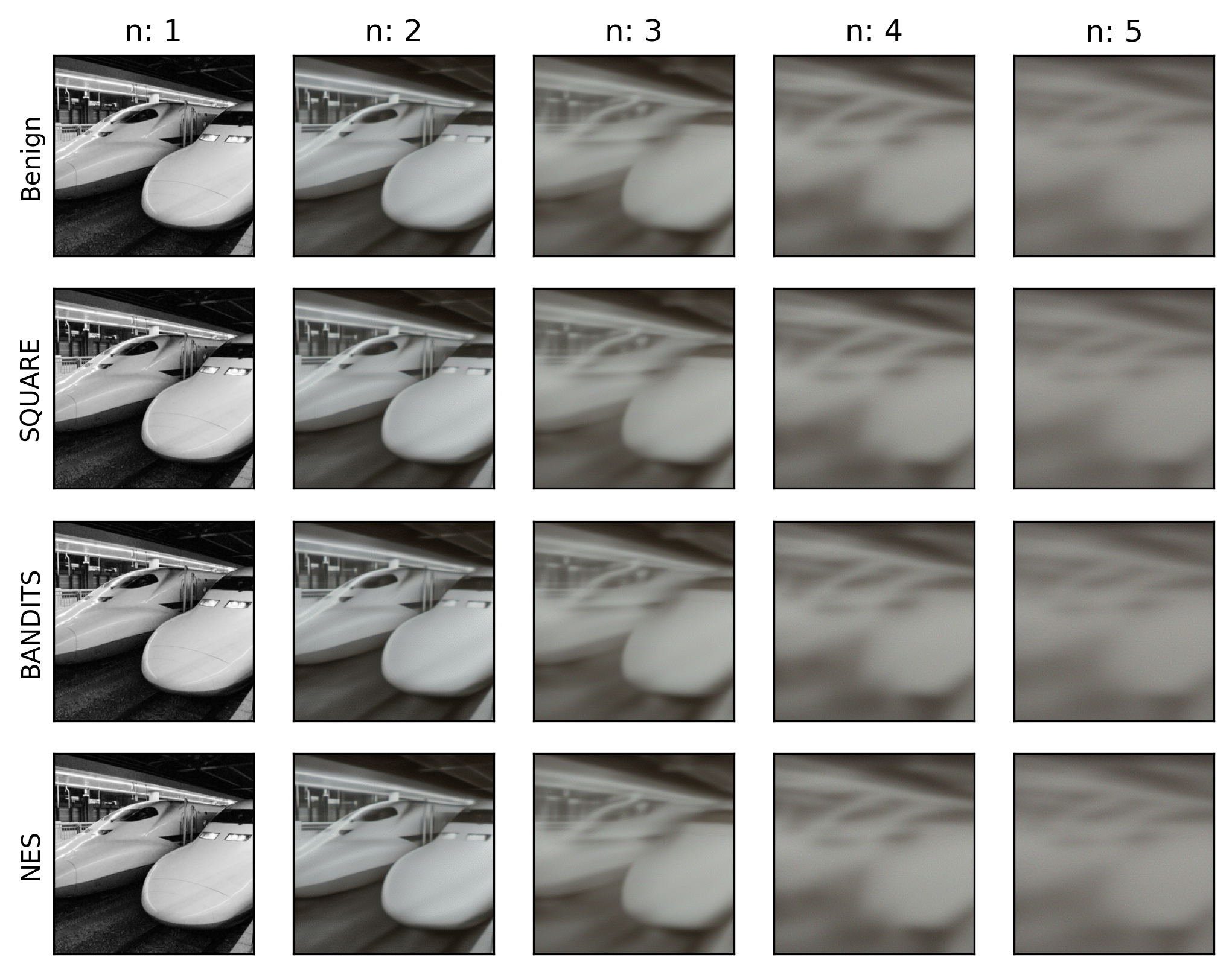}
    \caption{Recursive transformation from black-box attacked image (n: 1 to n: 5).}
    \label{fig:evolution_bb}
\end{figure}

\subsection{Transformations and Fourier Transformations} \label{app:evolution}
\begin{figure}[h]
    \centering
    \subfigure[n: 1. \label{fig:recursively}]{\includegraphics[width=0.45\textwidth]{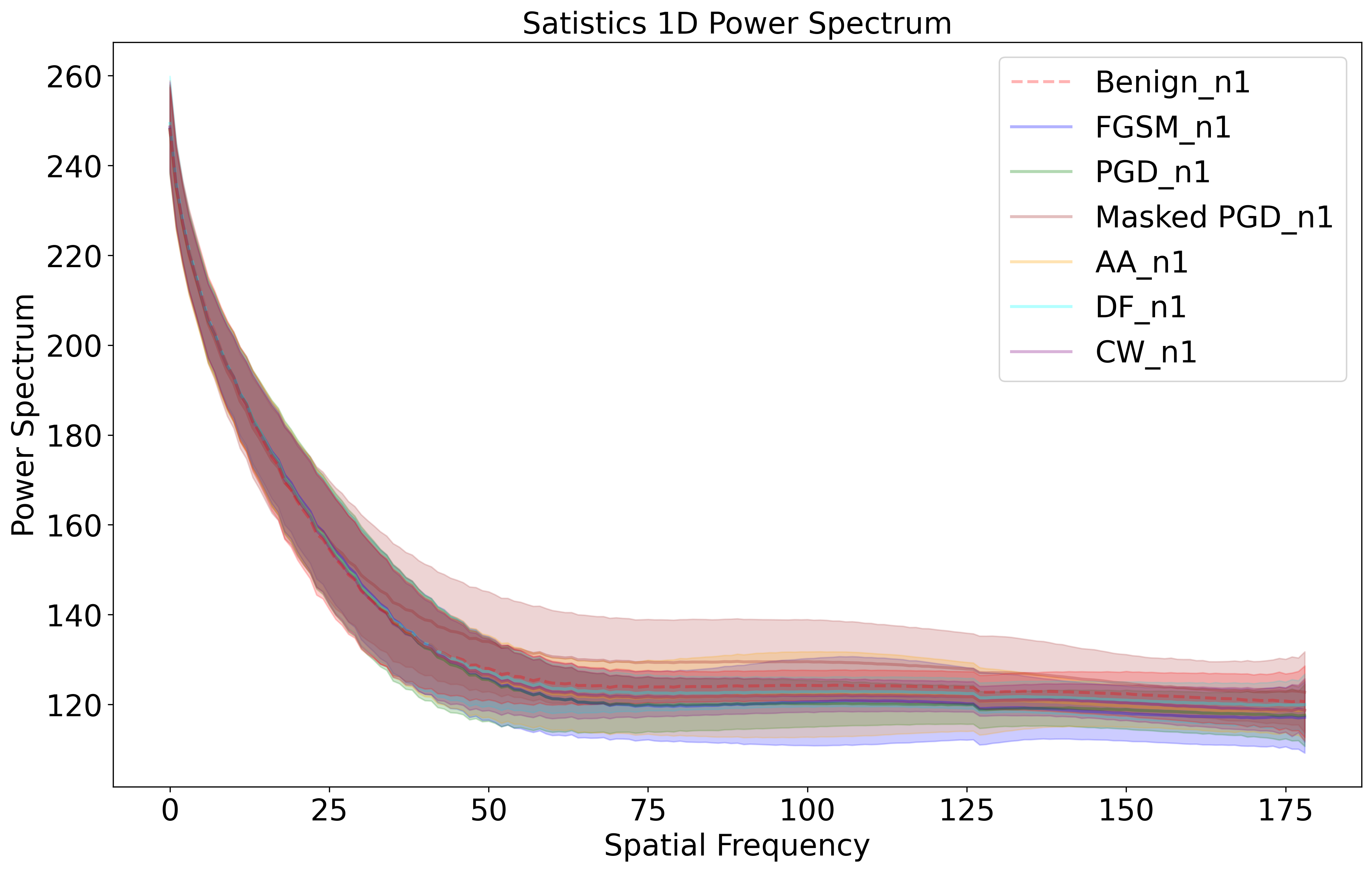} } \hfill
    \subfigure[n: 2.]{\includegraphics[width=0.45\textwidth]{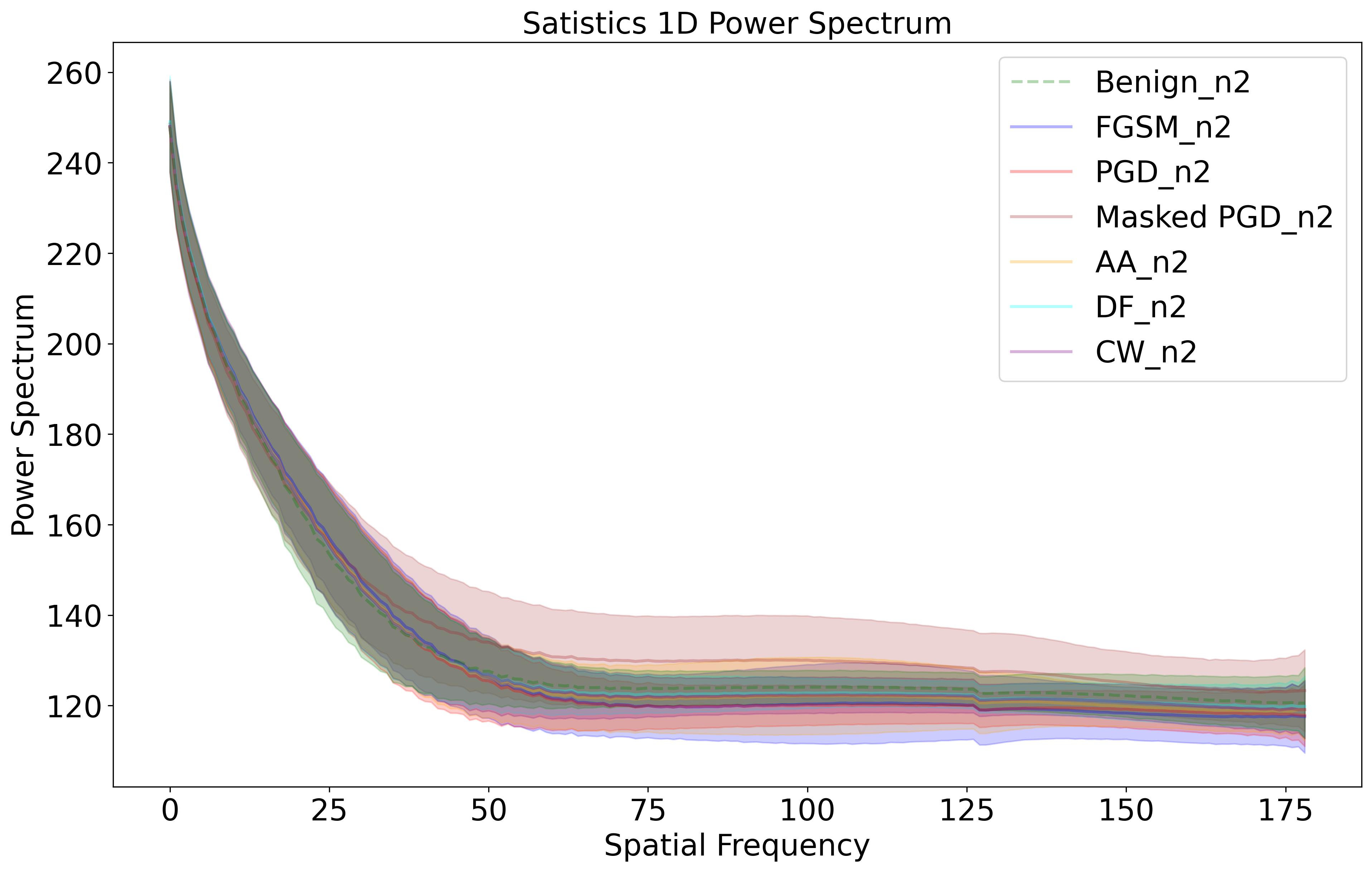}} 
    
    \subfigure[n: 3.]{\includegraphics[width=0.45\textwidth]{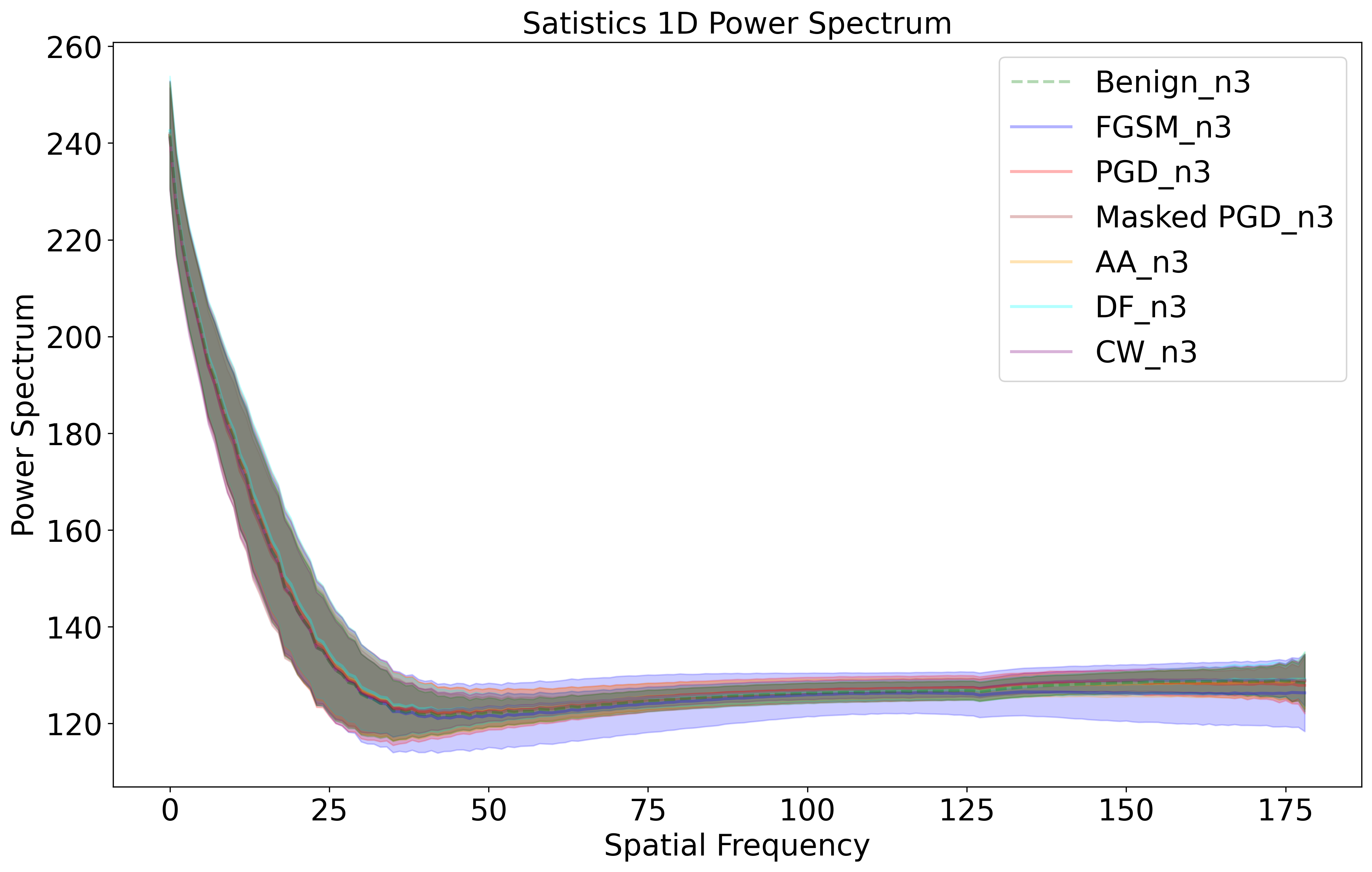}} \hfill
    \subfigure[n: 4.]{\includegraphics[width=0.45\textwidth]{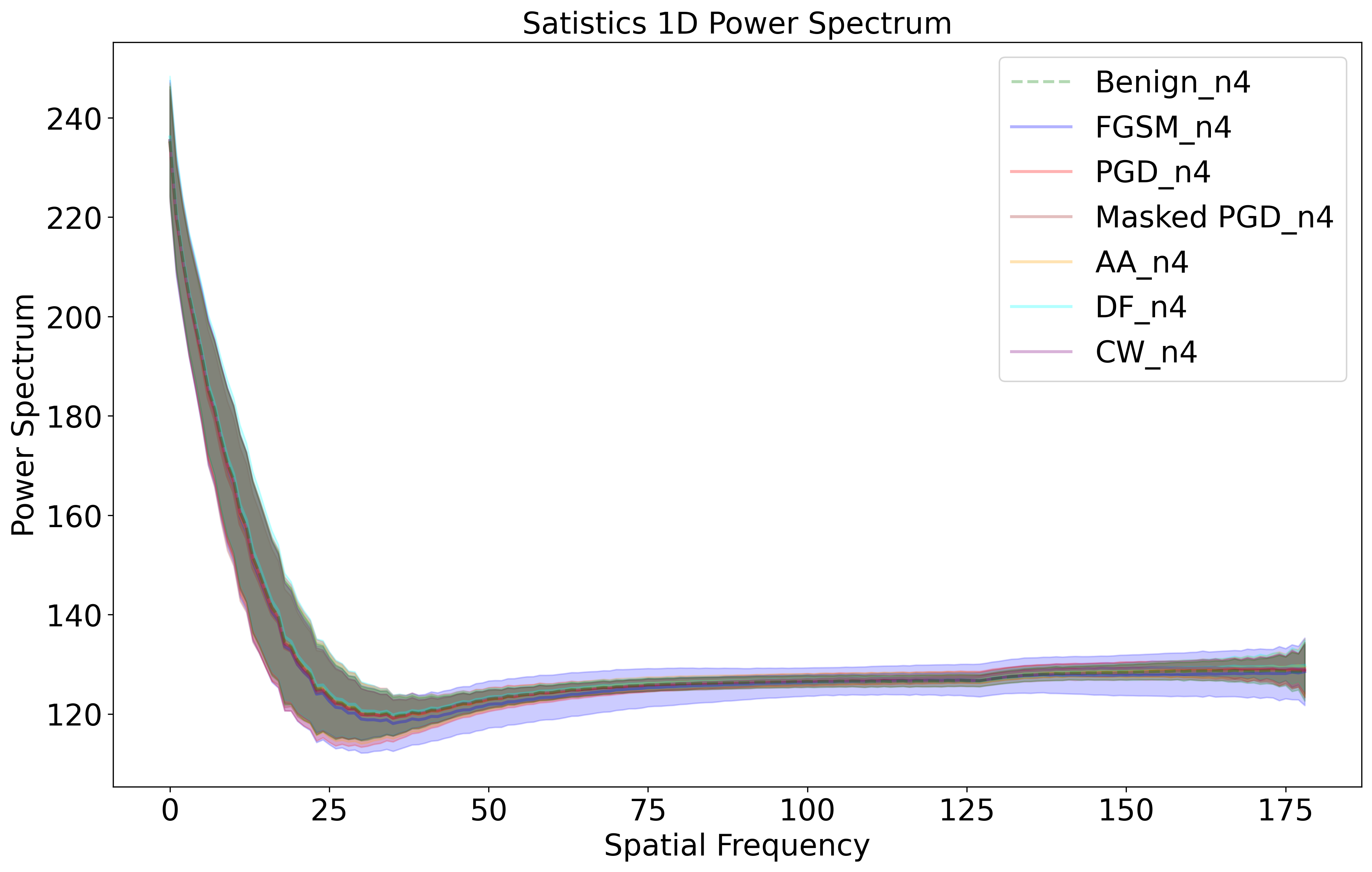}}
    
    \subfigure[n: 5.]{\includegraphics[width=0.45\textwidth]{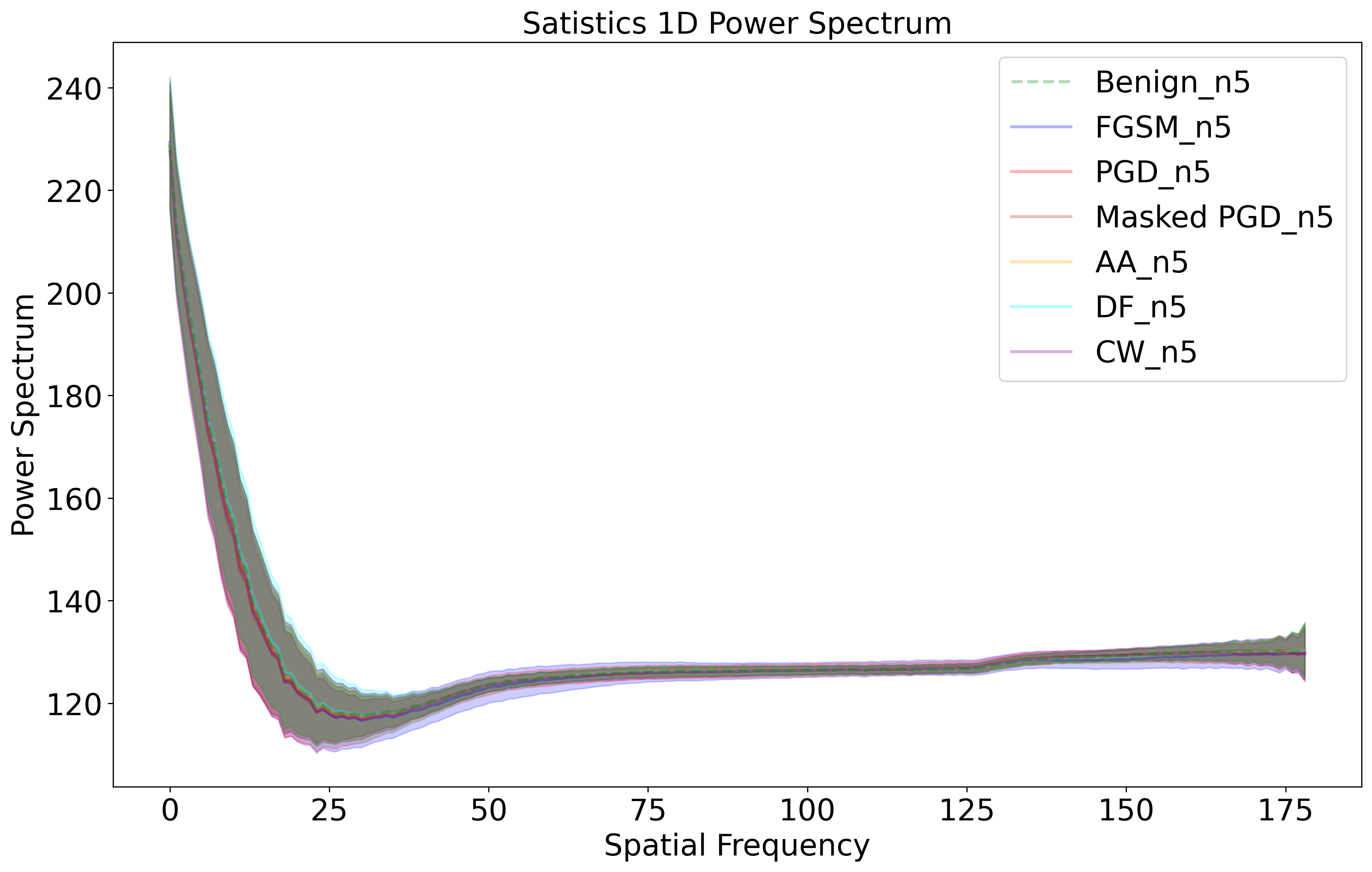}} \hfill
    \subfigure[n: 6.]{\includegraphics[width=0.45\textwidth]{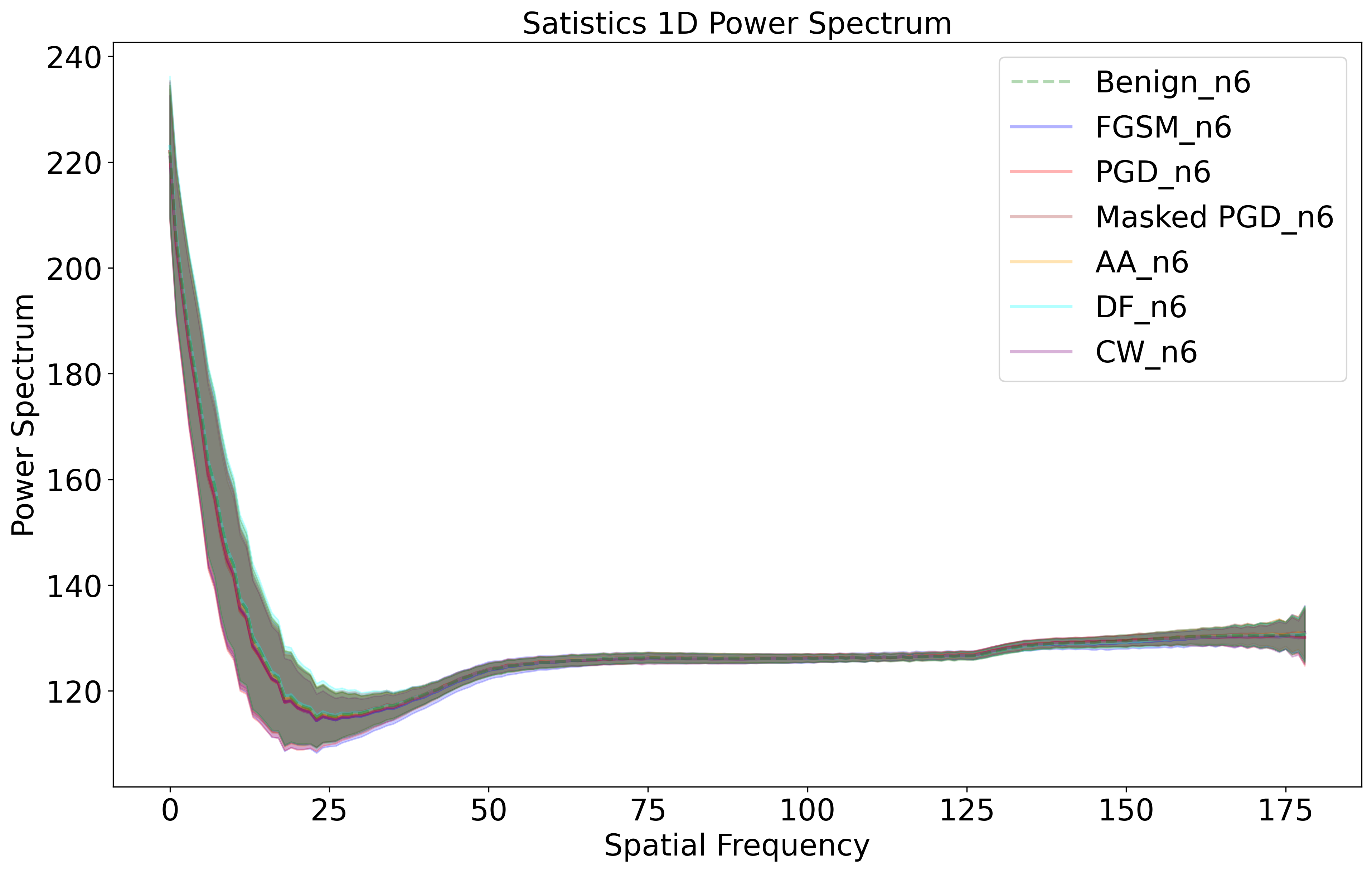}}
    \caption{
    1D power spectrum statistics from each sub-data ImageNet set for each attack method. 
    The more often the transformation is recursively applied, the more the power-spectrum is overlapping.
    }
    \label{fig:powerspectrum}
\end{figure}

\begin{figure}[h]
    \centering
    \includegraphics[width=0.5\textwidth]{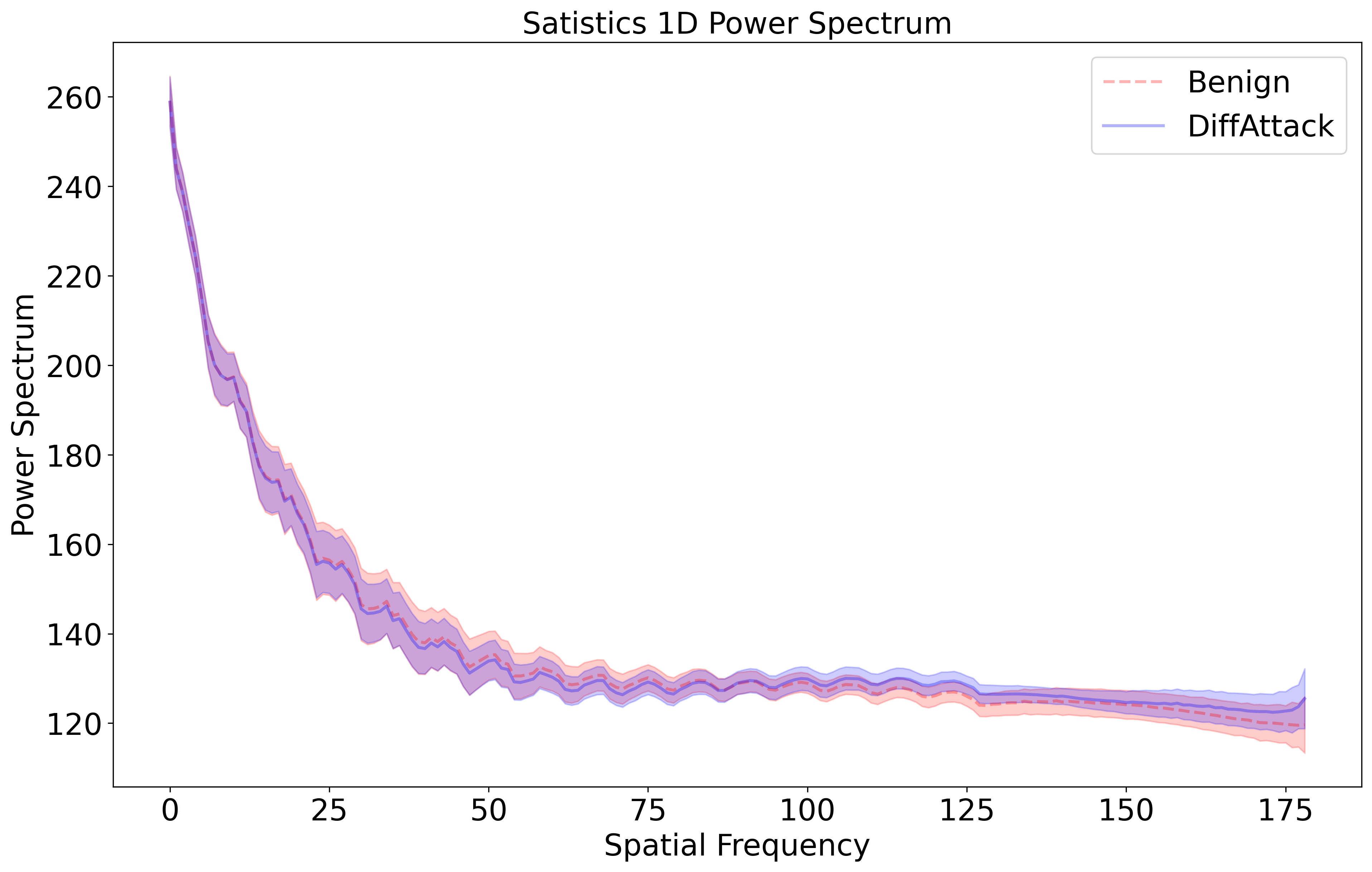}
    \caption{
    1D power spectrum statistics from each sub-data for 
    DiffAttack \cite{chen2023diffusion} on  ImageNet-Compressed dataset \cite{pytorch-playground}.
    }
    \label{fig:powerspectrum_imagenetcompatible}
\end{figure}

In this section, we investigate the reconstructed images by the Fourier transformation, i.e. \cref{fig:powerspectrum_imagenetcompatible} and  \cref{fig:powerspectrum}. 
Adversarial examples detection in supervised manners in the Fourier domain is already heuristic proven in \cite{lorenz2021detecting}.

We transform every sample from the spatial domain to the 1D frequency domain, reducing it to a 1D Power Spectrum. 
This method is formed by a Discrete Fourier Transform followed by an azimuthally average.
The transformation can be substantially optimized by employing the Fast Fourier Transform.
Notice that after applying the transformation, we use only the power spectrum.
A small shift after the first transformation can be recognized, as shown in \cref{fig:recursively}.
Therefore, we apply the inverse and reverse process several times and analyze it after each transformation with the FFT analysis.
To this end, all FFT spectrums almost totally overlap after applying the recursive transformation process several times.

\subsection{Transferability Capabilities} \label{app:transferability}

\begin{figure}[h]
    \centering
    \subfigure[8/255.]{\includegraphics[width=0.45\textwidth]{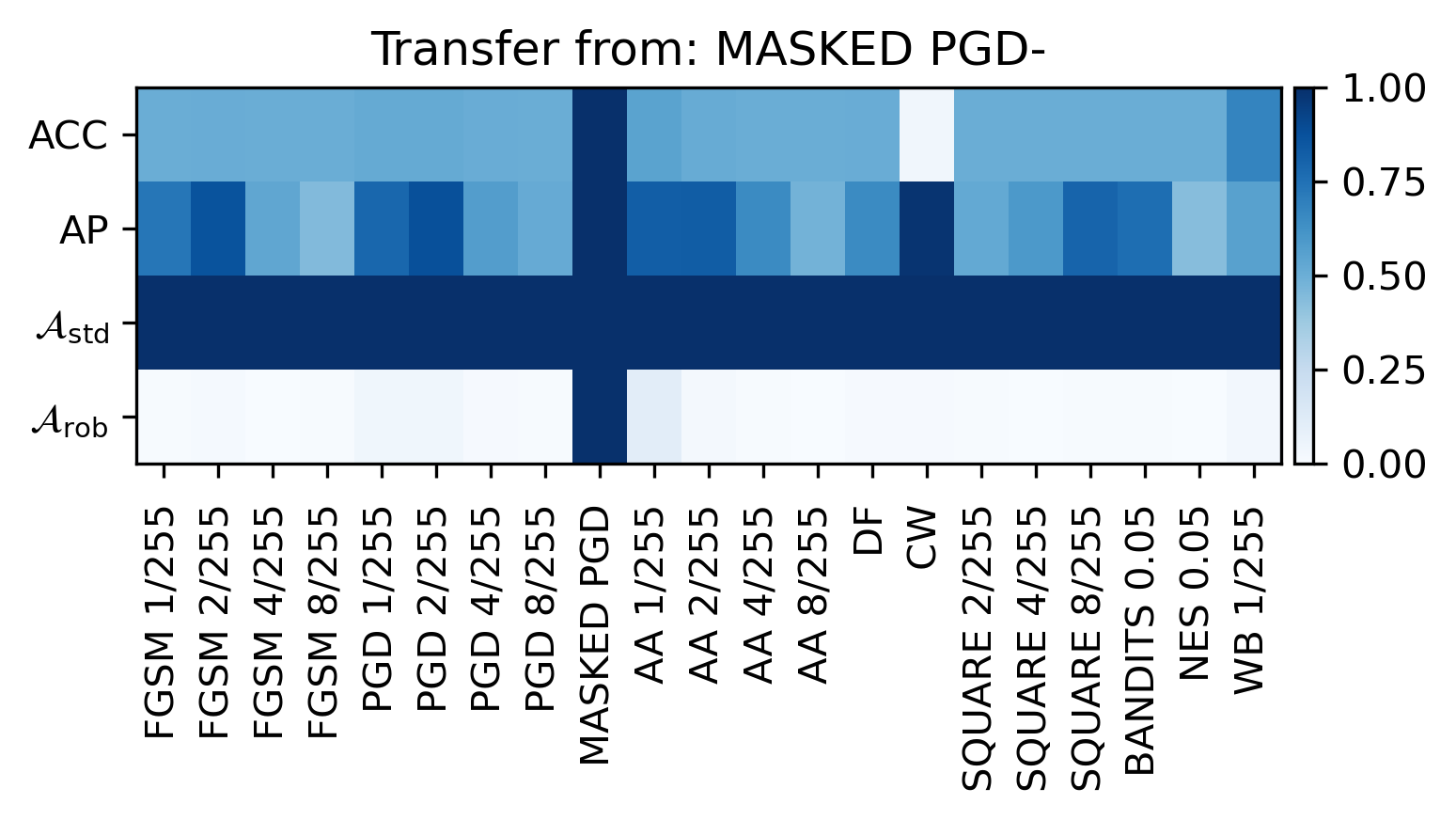}} \hfill
    \subfigure[4/255.]{\includegraphics[width=0.45\textwidth]{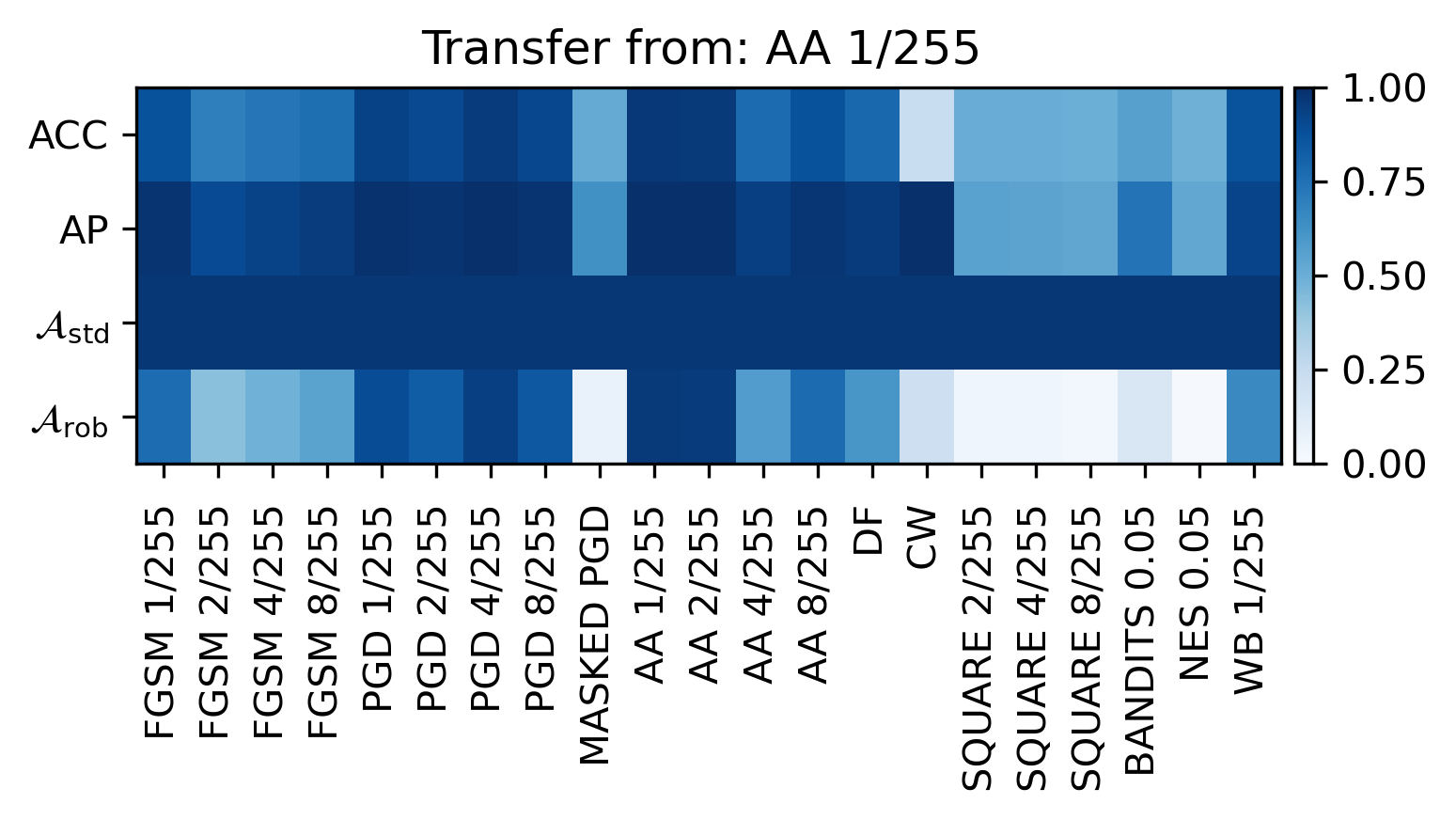}}
    
    \subfigure[2/255.]{\includegraphics[width=0.45\textwidth]{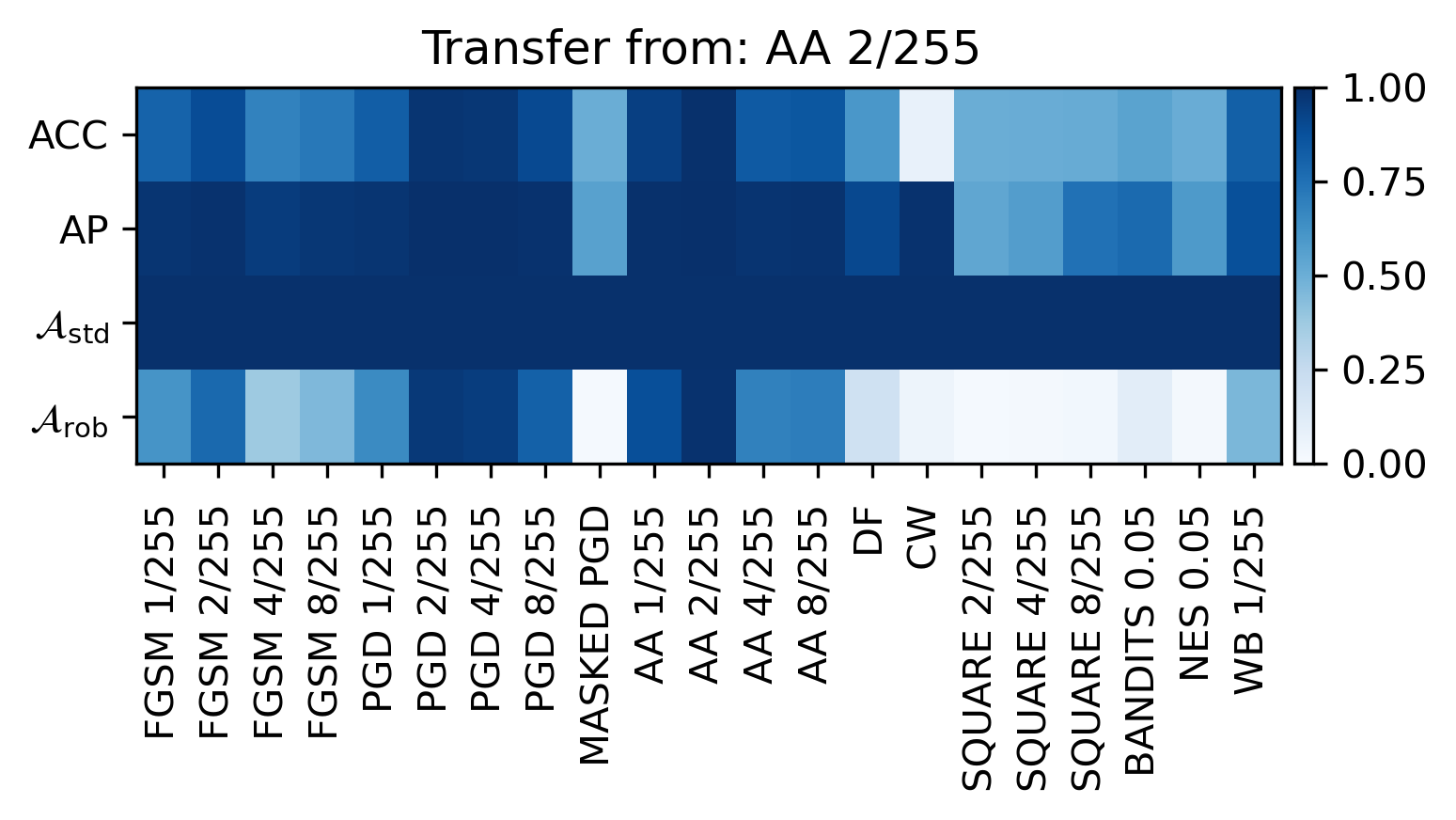}} \hfill
    \subfigure[1/255.]{\includegraphics[width=0.45\textwidth]{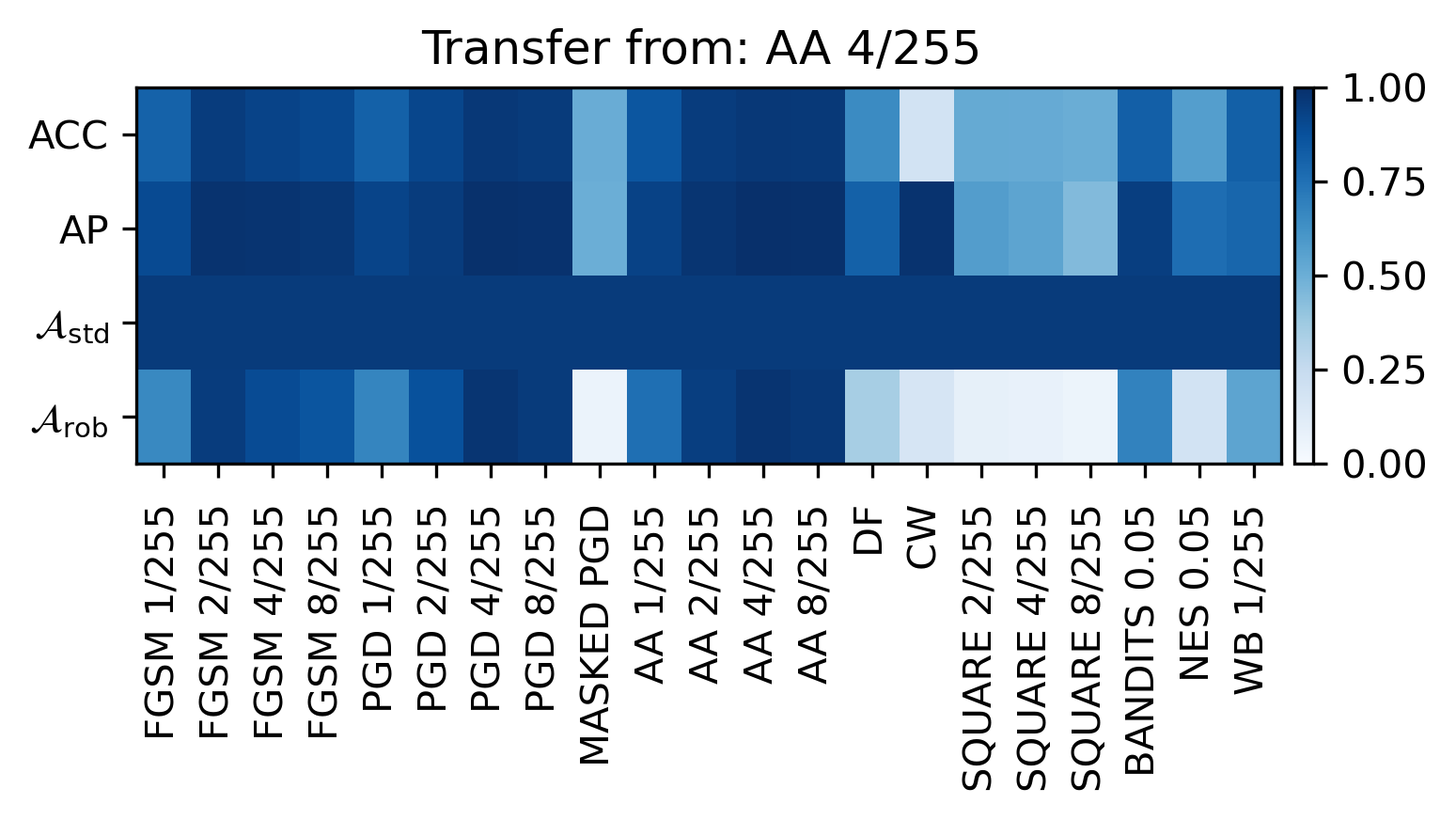}}
    
    \caption{Transferability. From FGSM to the other attacks.}
    \label{fig:transfer_fgsm}
\end{figure}

\begin{figure}[h]
    \centering
    \subfigure[8/255.]{\includegraphics[width=0.45\textwidth]{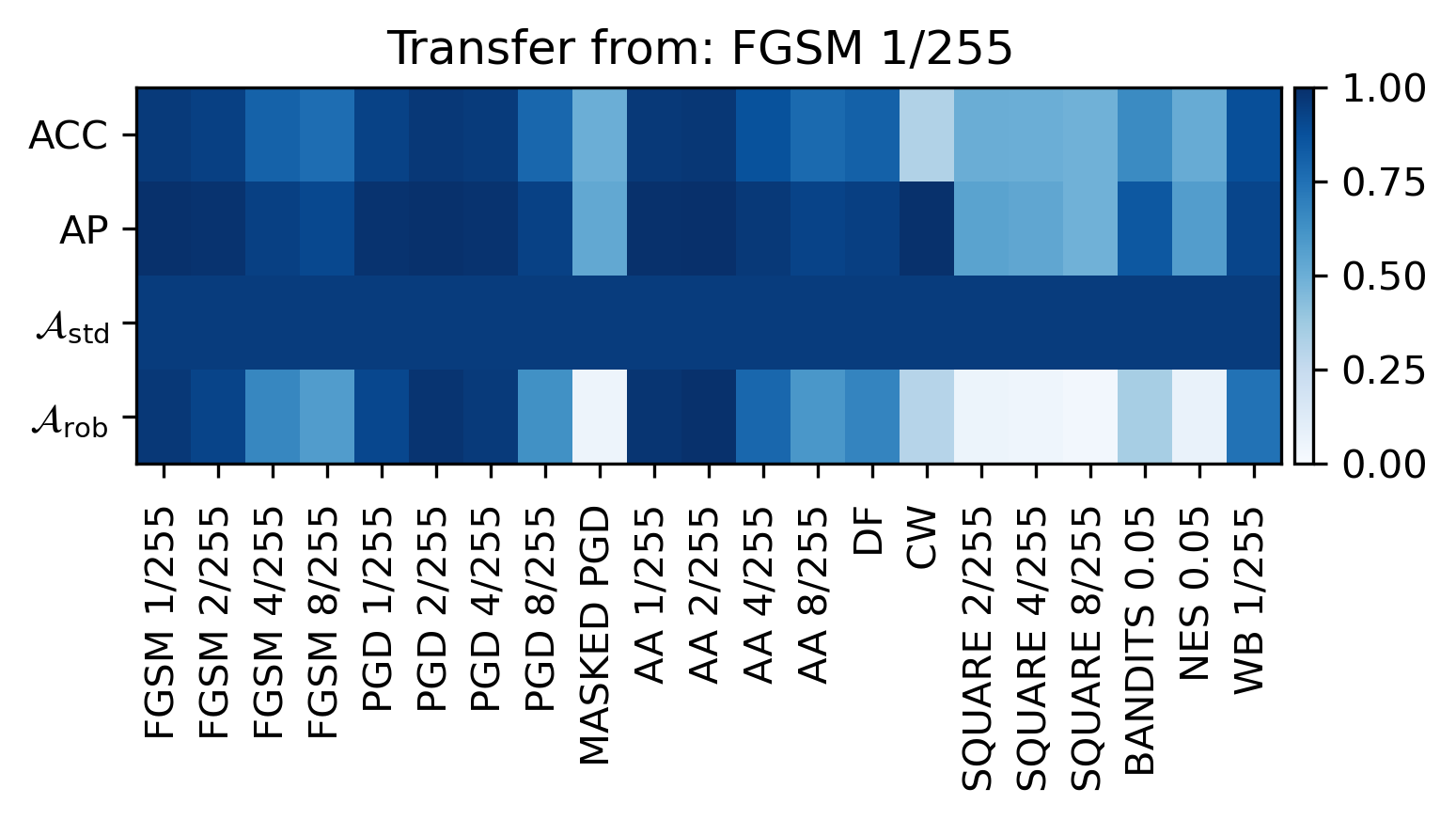}} \hfill
    \subfigure[4/255.]{\includegraphics[width=0.45\textwidth]{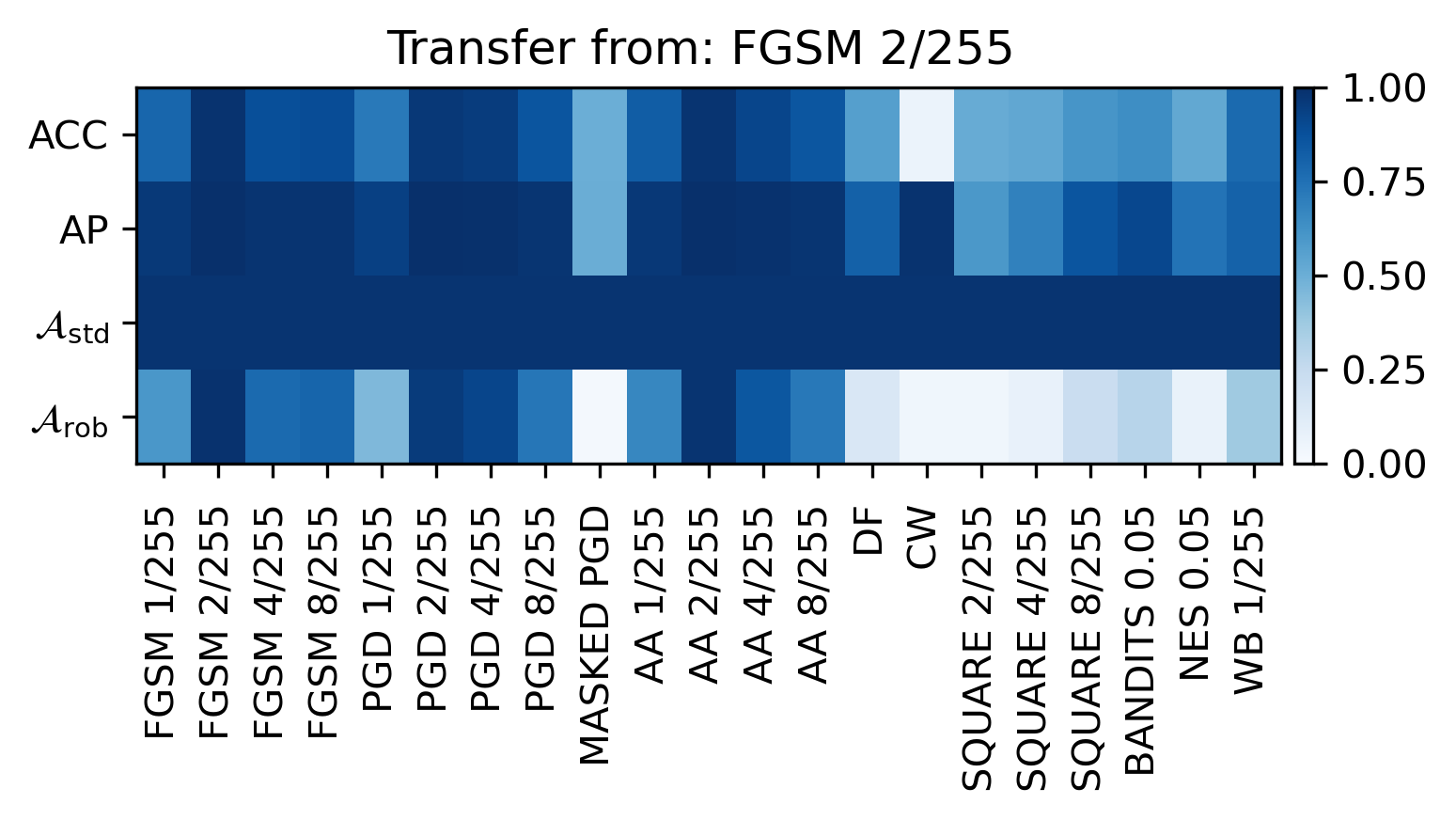}}
    
    \subfigure[2/255.]{\includegraphics[width=0.45\textwidth]{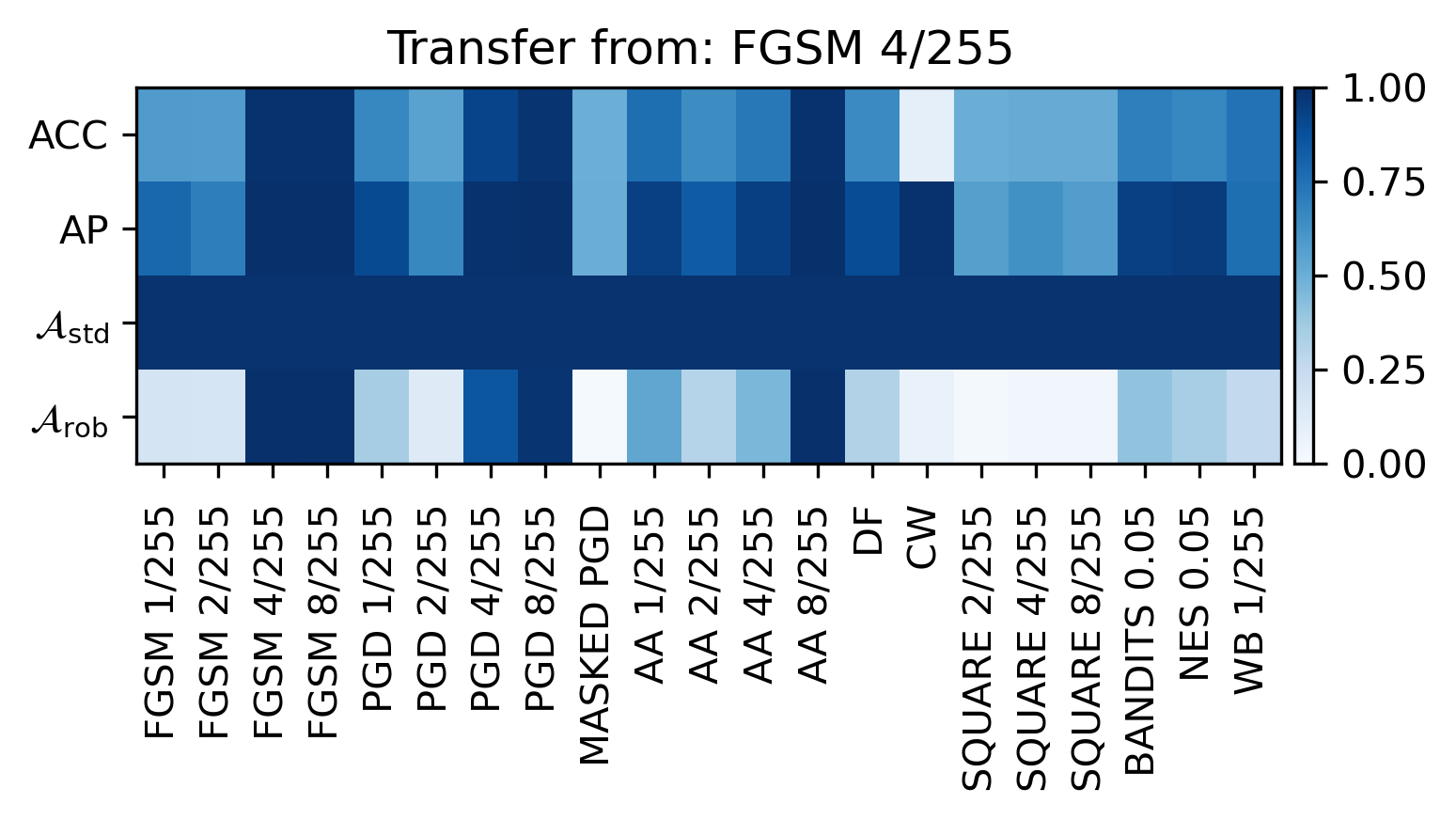}} \hfill
    \subfigure[1/255.]{\includegraphics[width=0.45\textwidth]{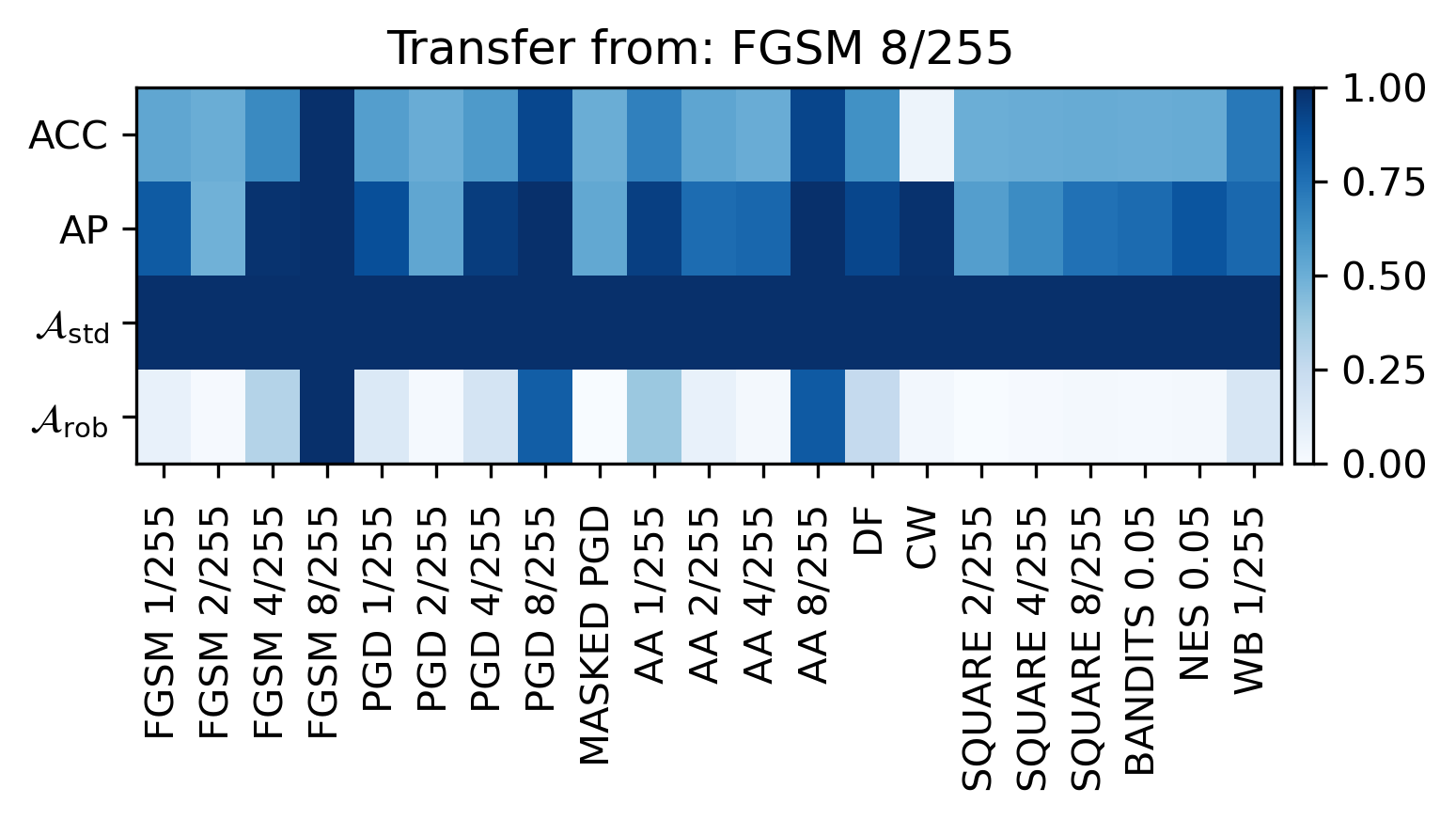}}
    
    \caption{Transferability. From PGD to the other attacks.}
    \label{fig:transfer_pgd}
\end{figure}

\begin{figure}[h]
    \centering
    \includegraphics[width=0.45\textwidth]{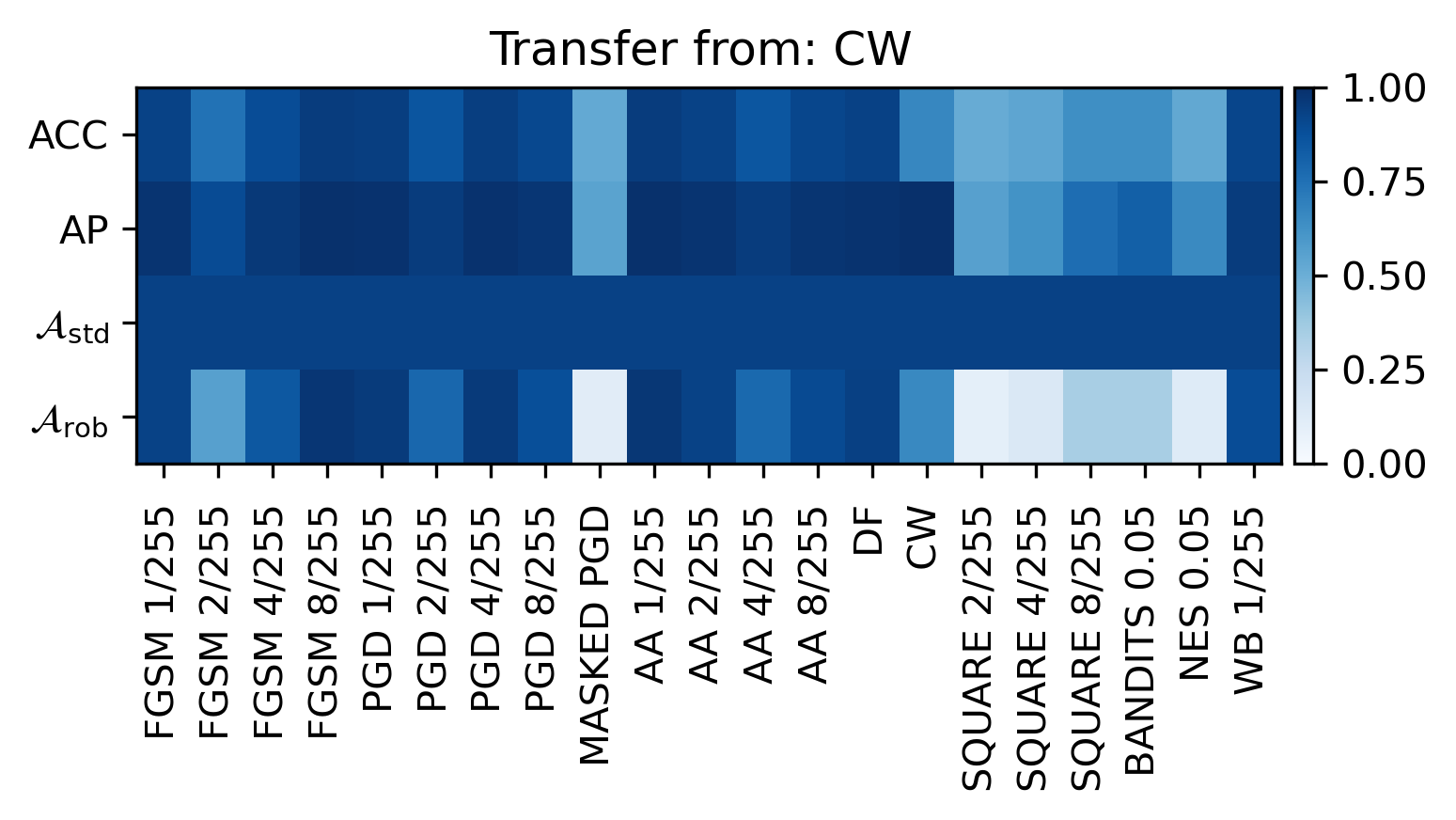}
    \caption{Transferability. From Masked PGD to the other attacks.}
    \label{fig:transfer_mPGD}
\end{figure}

\begin{figure}[h]
    \centering
    \subfigure[8/255.]{\includegraphics[width=0.45\textwidth]{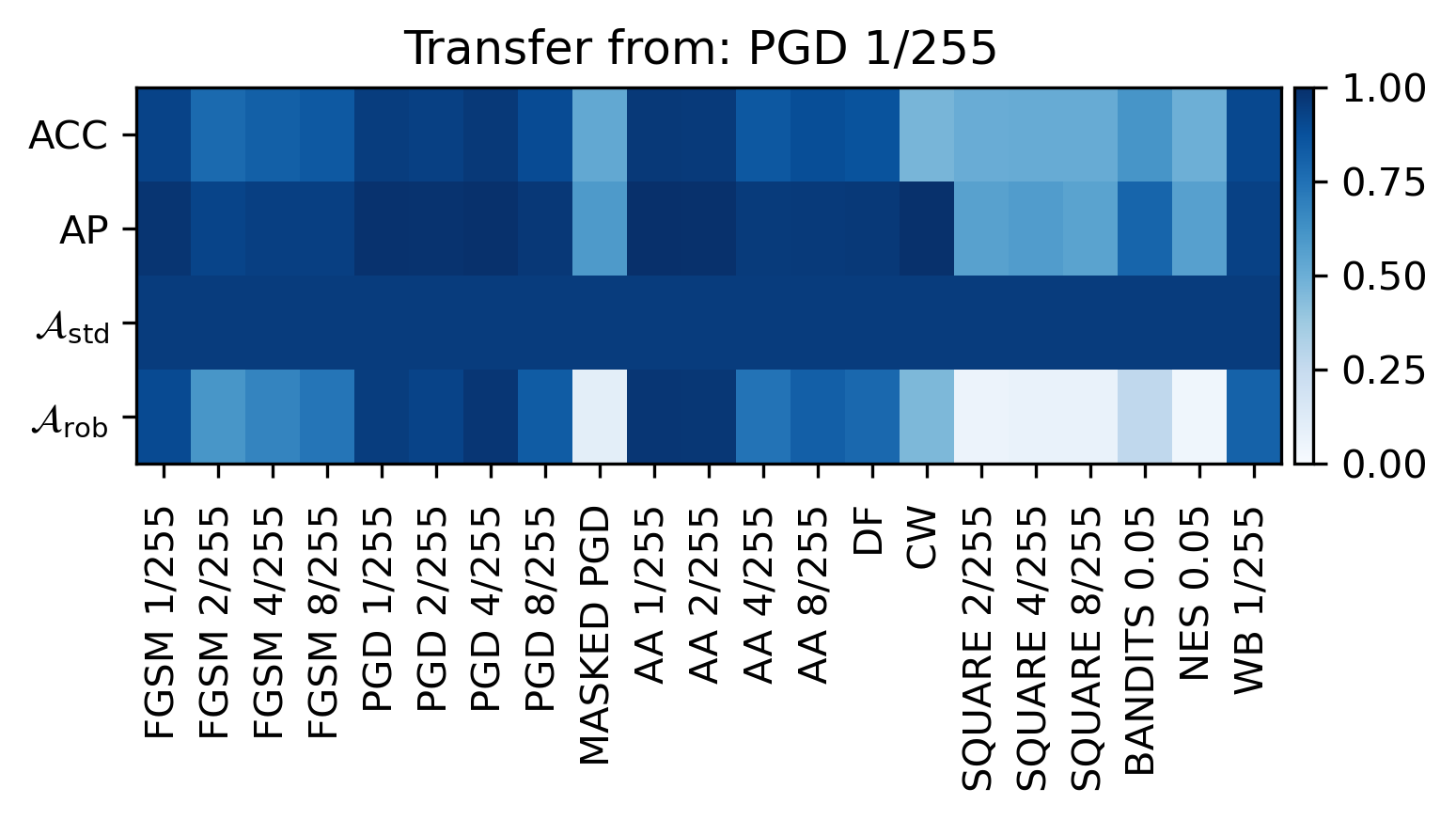}} \hfill
    \subfigure[4/255.]{\includegraphics[width=0.45\textwidth]{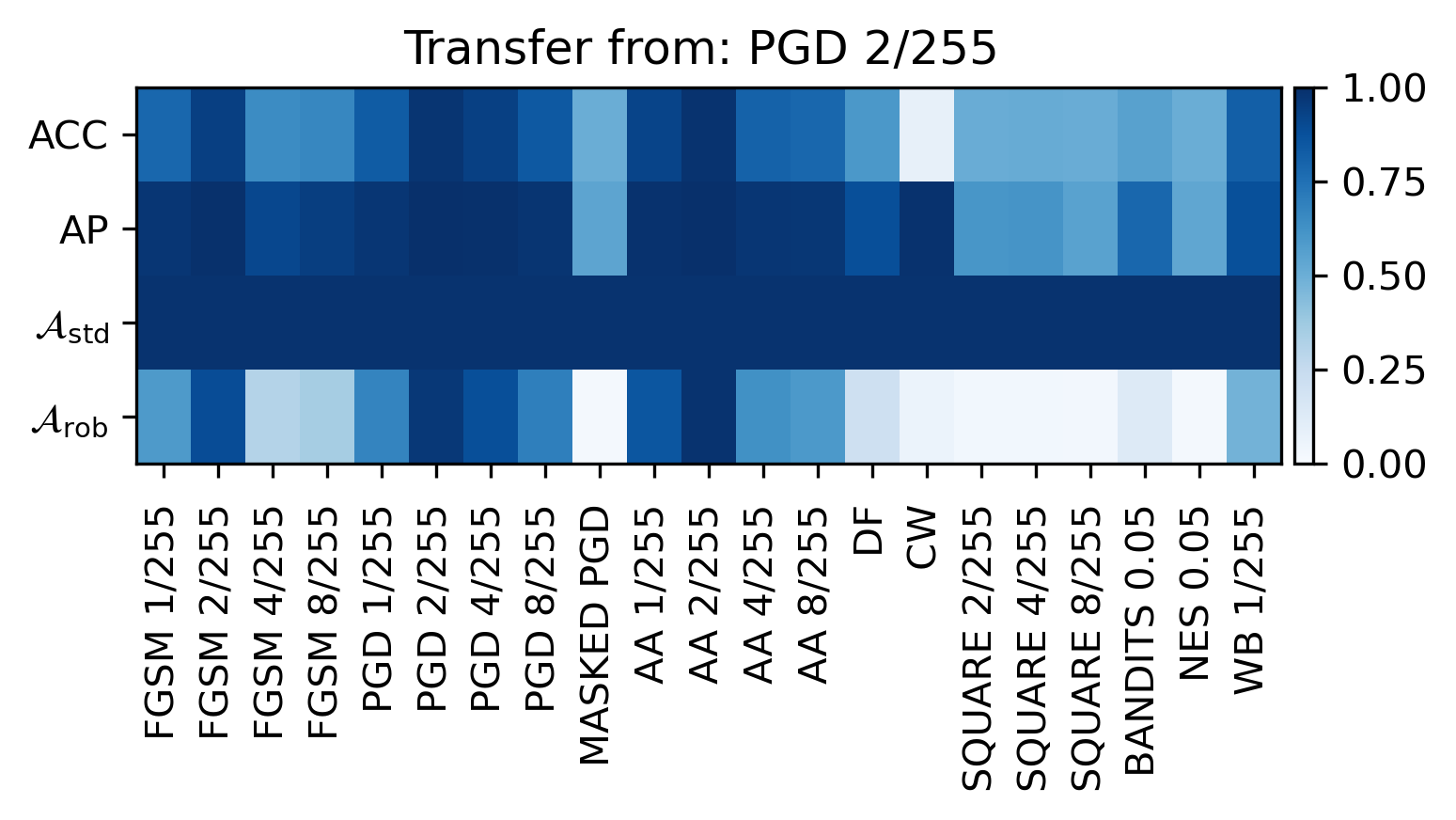}}
    
    \subfigure[2/255.]{\includegraphics[width=0.45\textwidth]{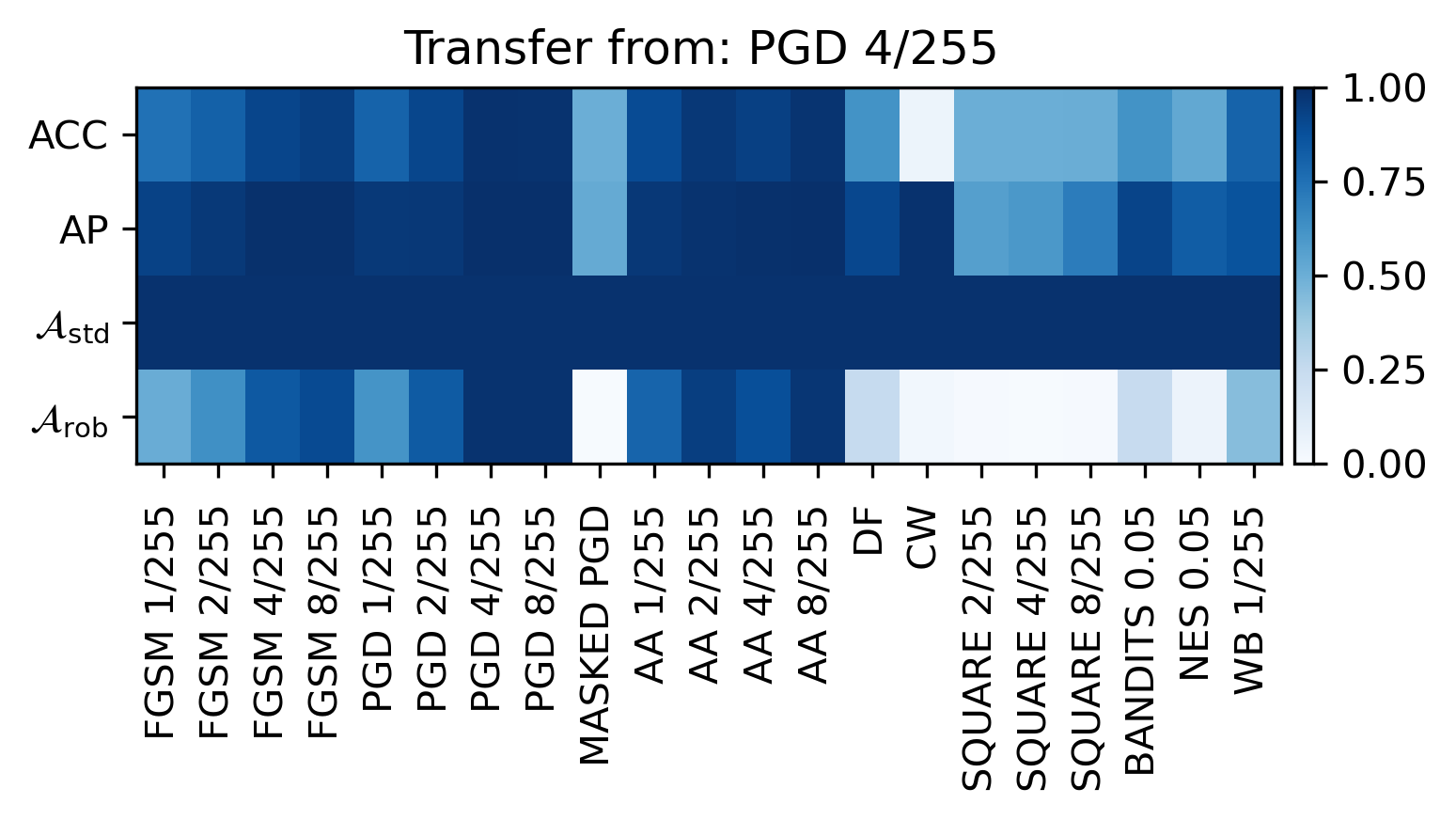}} \hfill
    \subfigure[1/255.]{\includegraphics[width=0.45\textwidth]{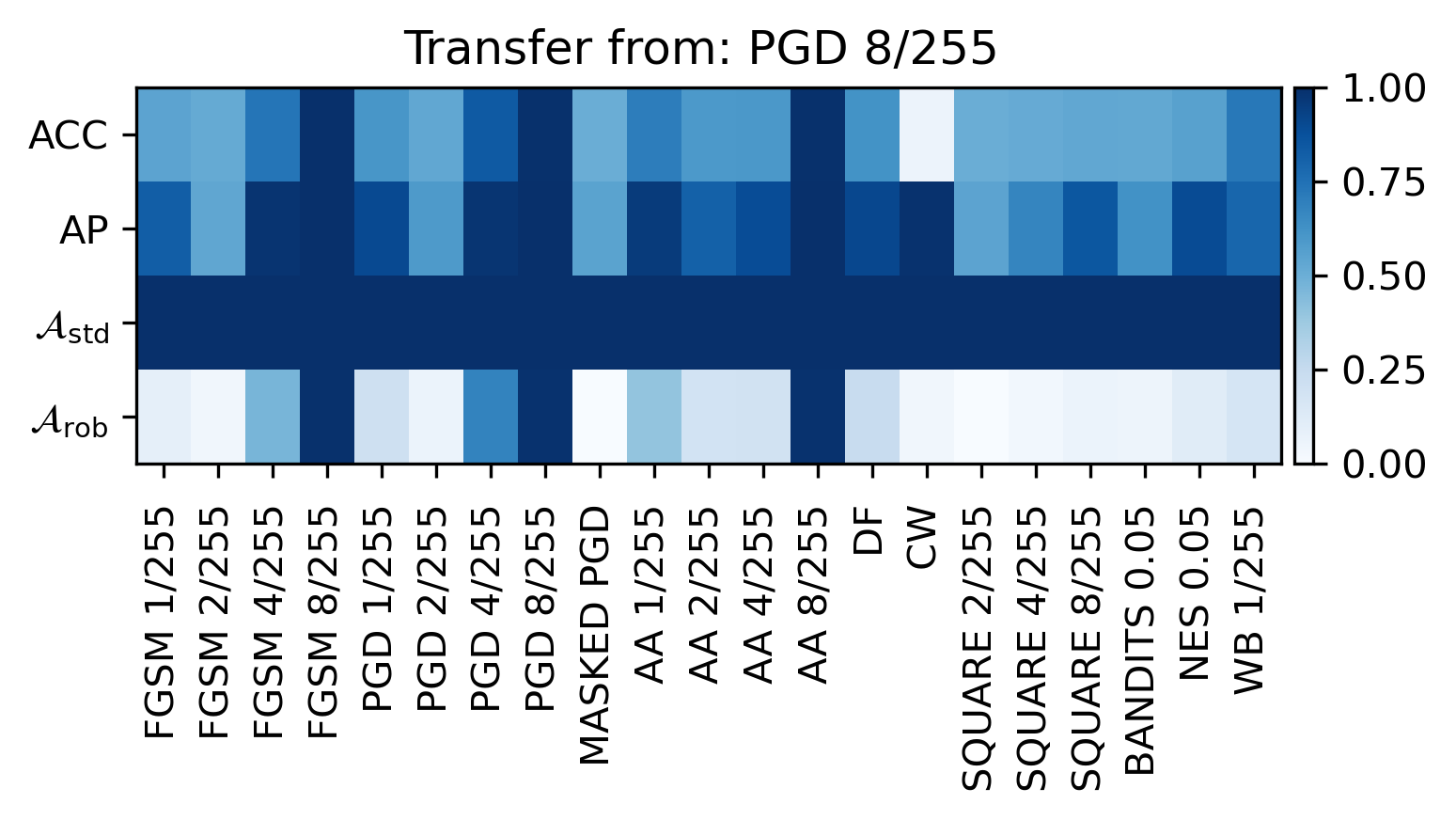}}
    
    \caption{Transferability. From AA to the other attacks.}
    \label{fig:transfer_aa}
\end{figure}

\begin{figure}[h]
\centering
    \begin{minipage}[b]{0.45\textwidth}
        \centering
        \includegraphics[width=\linewidth]{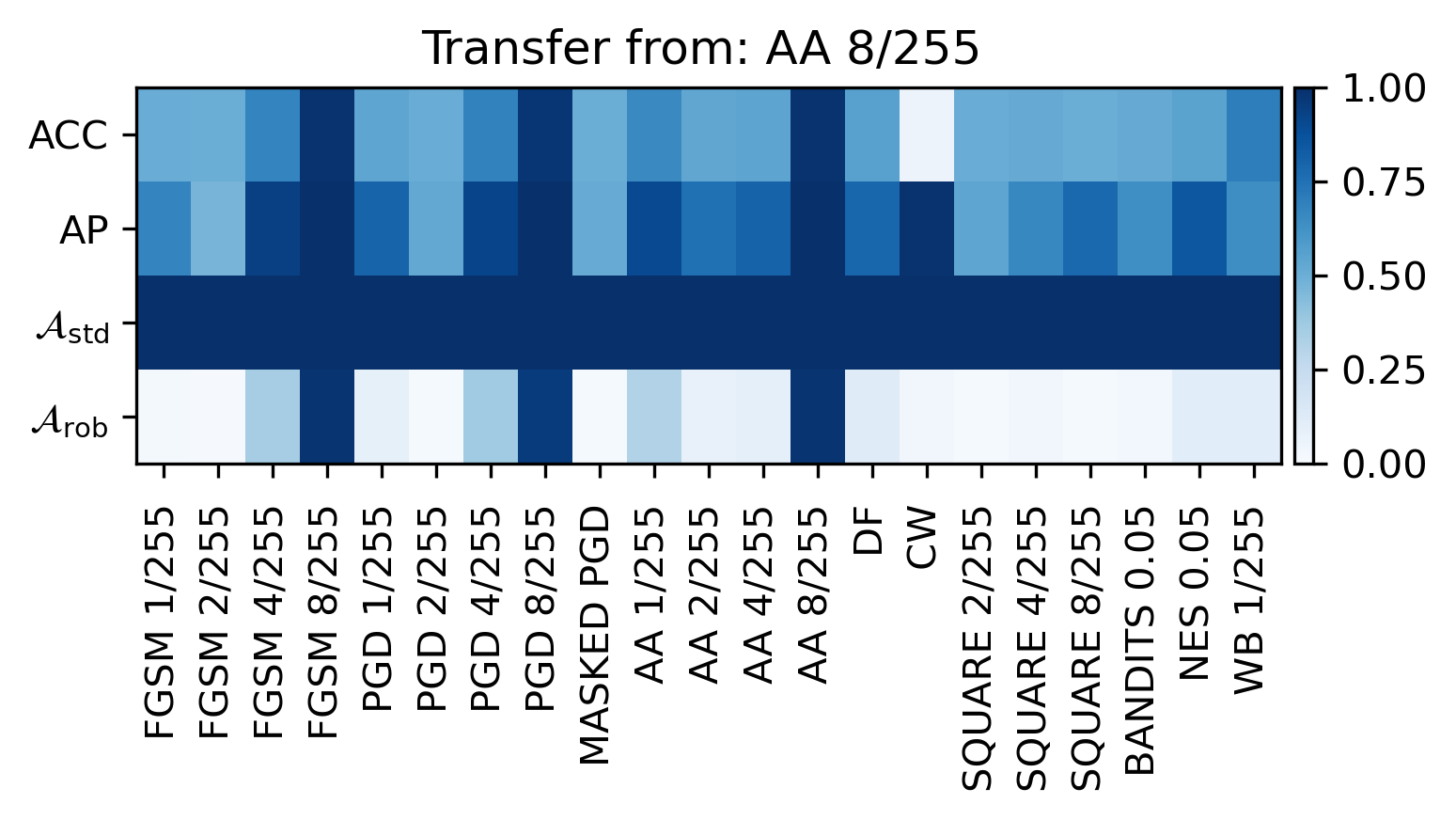}
        \caption{Transferability. From CW to the other attacks.}
        \label{fig:transfer_cw}
    \end{minipage}
    \hfill
    \begin{minipage}[b]{0.45\textwidth}
        \centering
        \includegraphics[width=\linewidth]{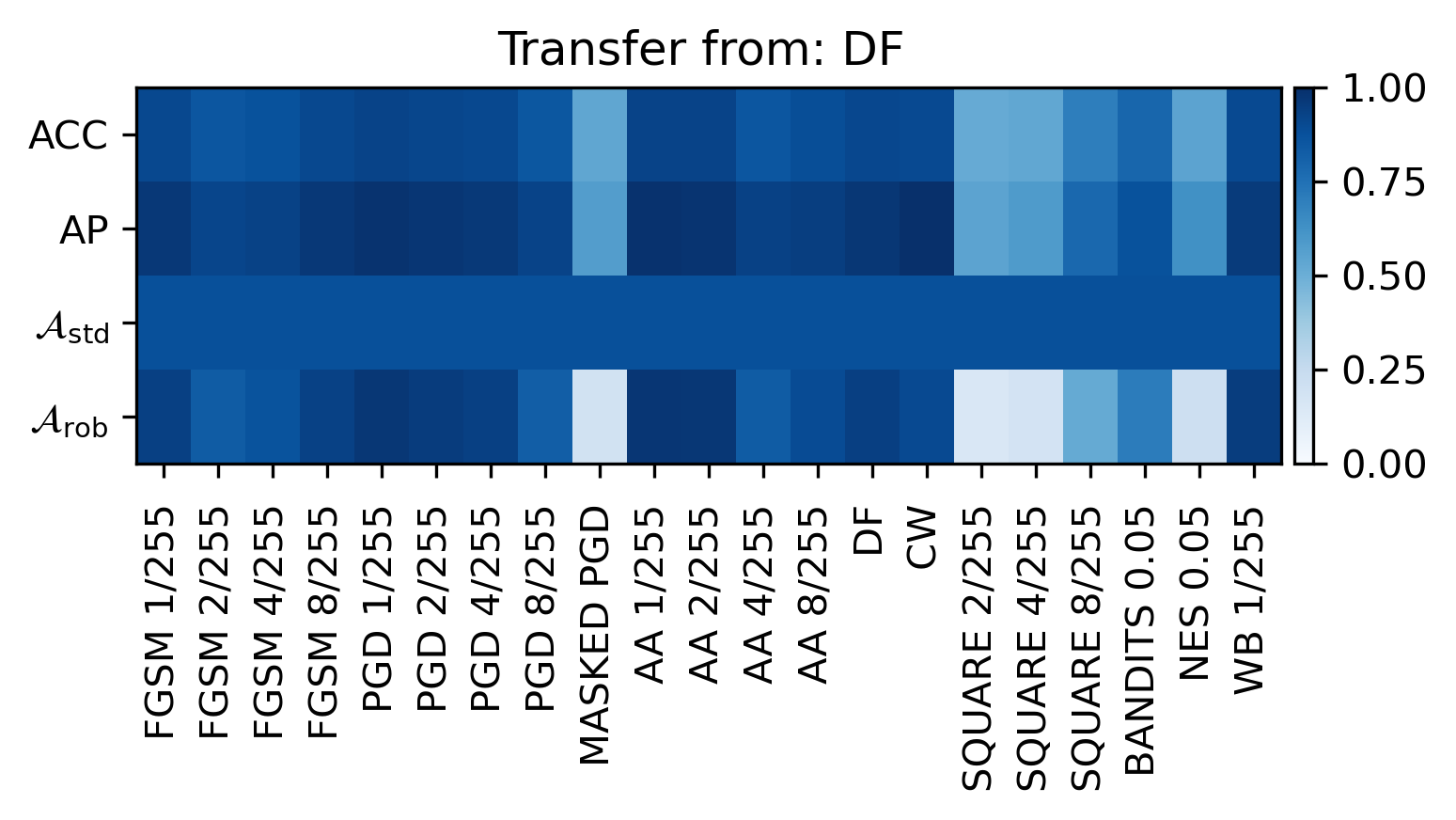}
        \caption{Transferability. From DF to the other attacks.}
        \label{fig:transfer_df}
    \end{minipage}
\end{figure}

\begin{figure}[h]
    \centering
    \subfigure[8/255.]{\includegraphics[width=0.45\textwidth]{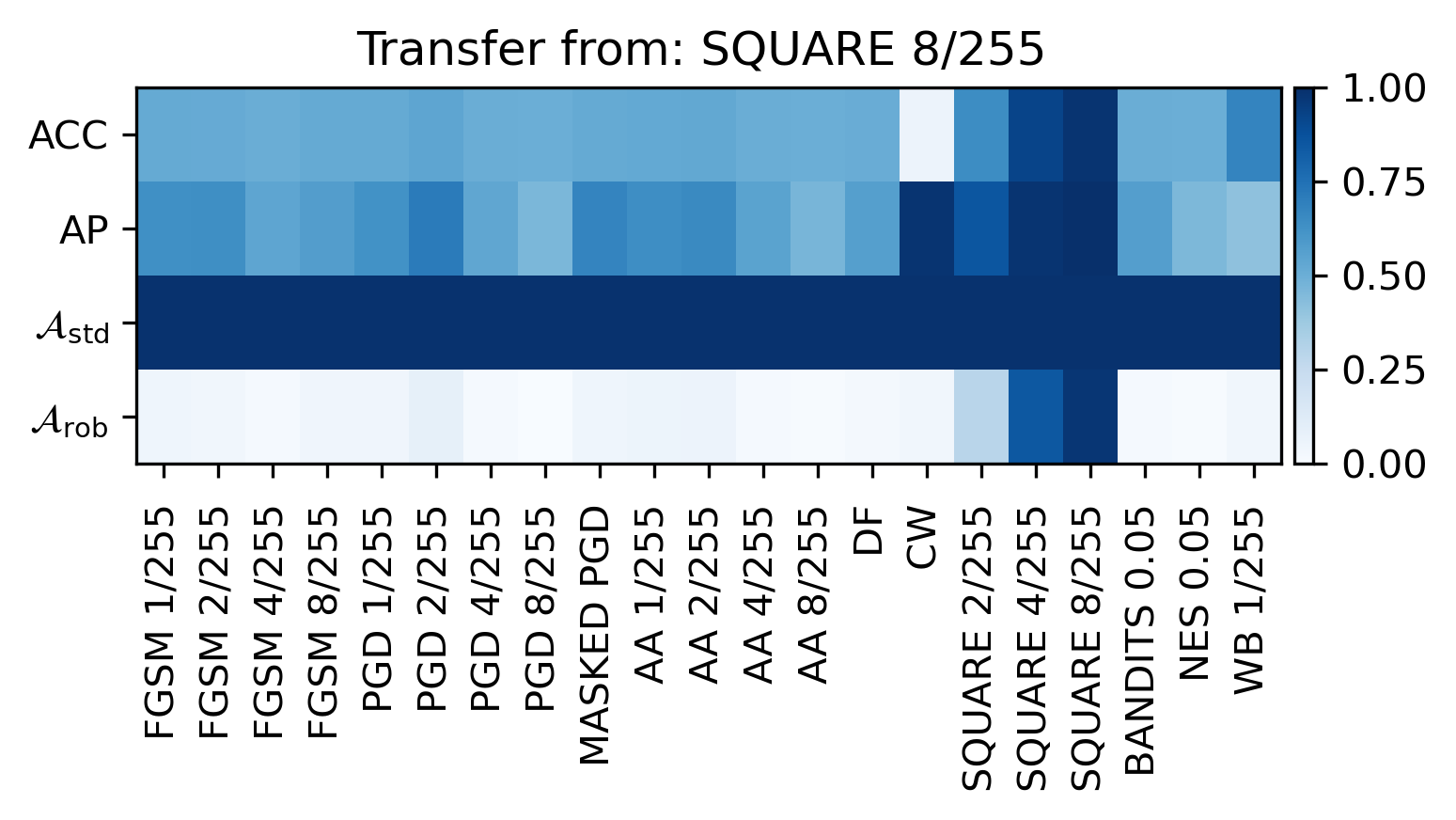}} \hfill
    \subfigure[4/255.]{\includegraphics[width=0.45\textwidth]{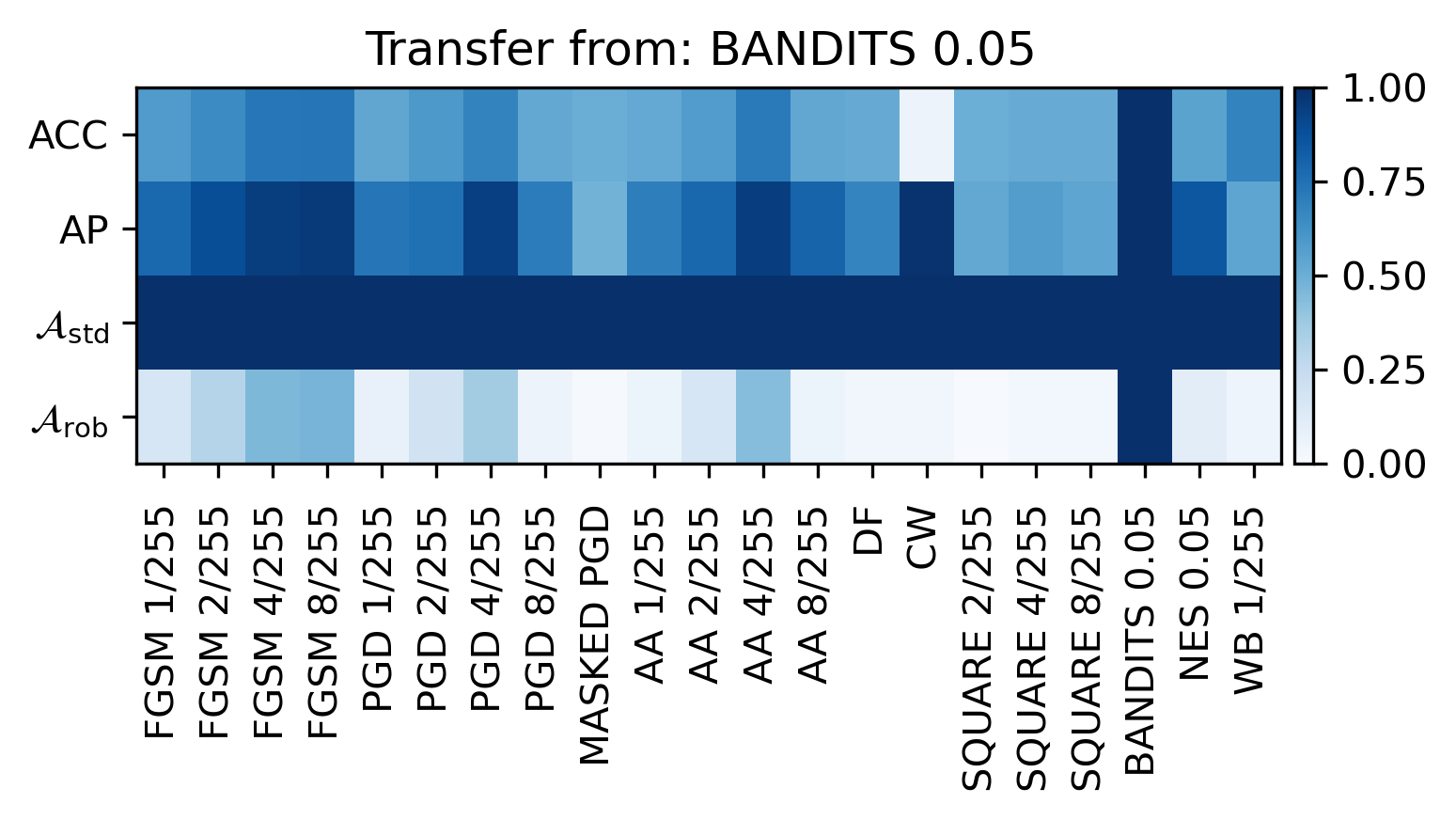}}
    
    \subfigure[2/255.]{\includegraphics[width=0.45\textwidth]{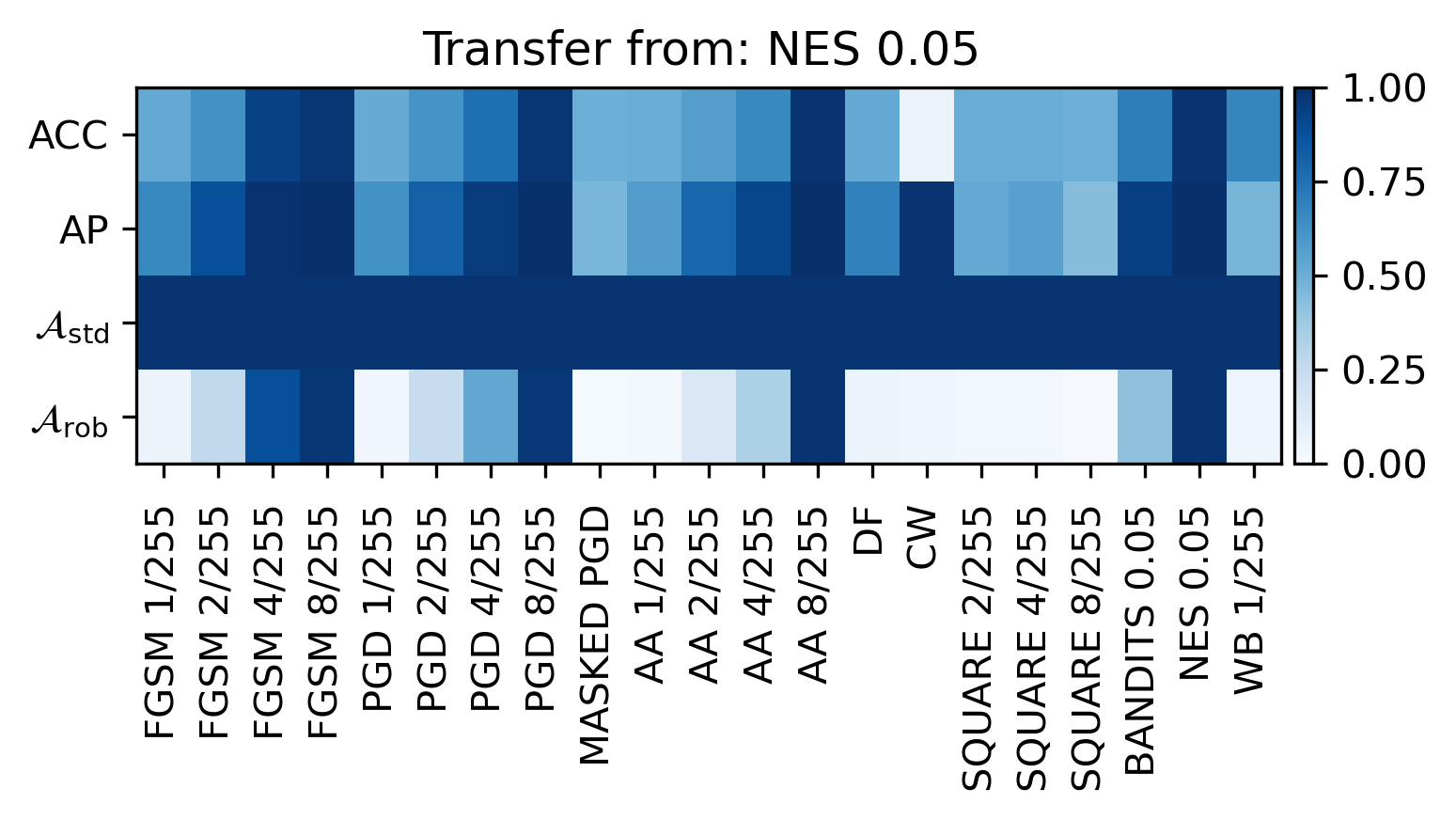}} \hfill
    
    \caption{Transferability. From Square to the other attacks.}
    \label{fig:transfer_square}
\end{figure}

\begin{figure}[h]
\centering
    \begin{minipage}[b]{0.45\textwidth}
        \centering
        \includegraphics[width=\linewidth]{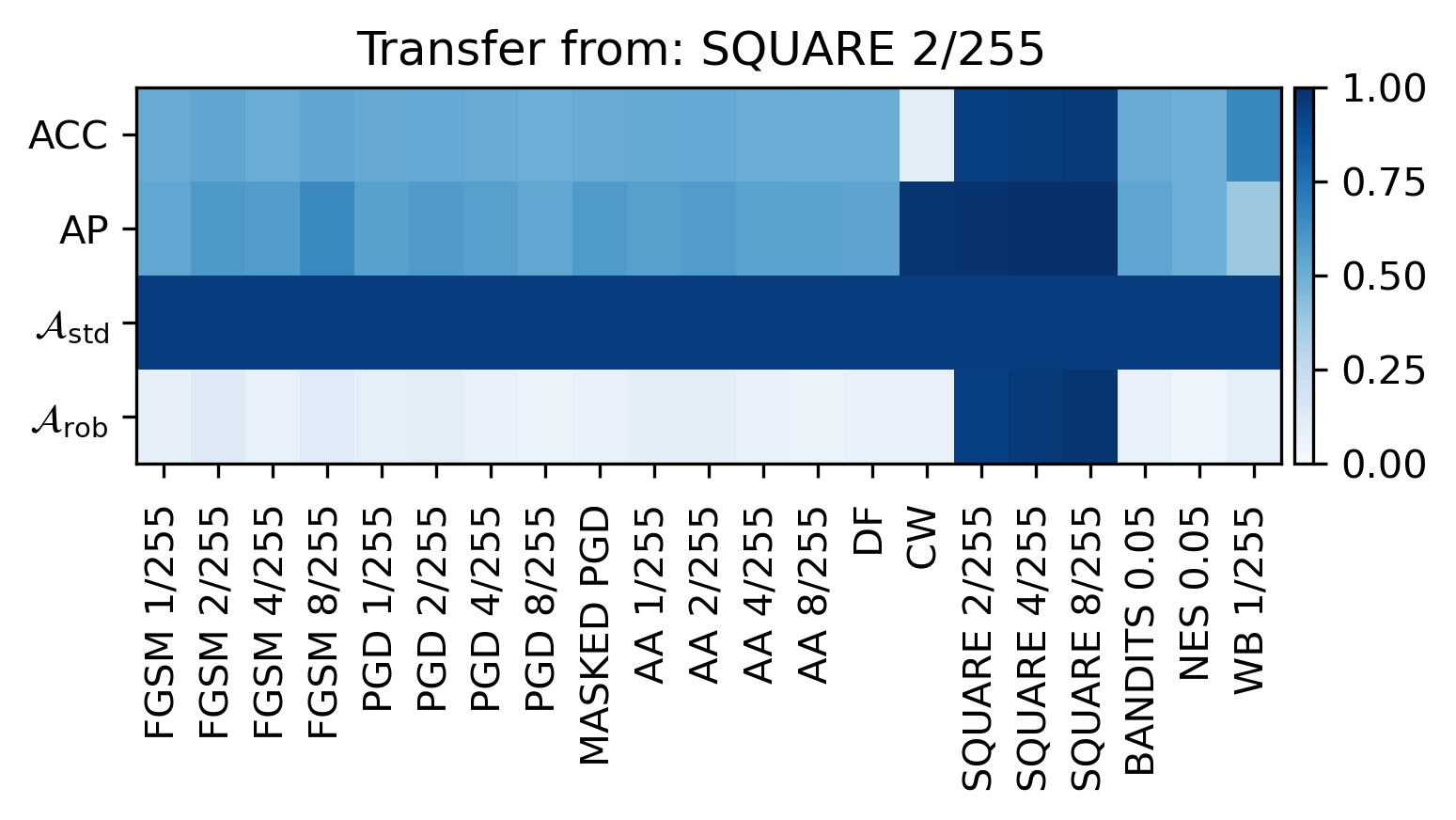}
        \caption{Transferability. From Bandits to the other attacks.}
        \label{fig:transfer_bandits}
    \end{minipage}
    \hfill
    \begin{minipage}[b]{0.45\textwidth}
        \centering
        \includegraphics[width=\linewidth]{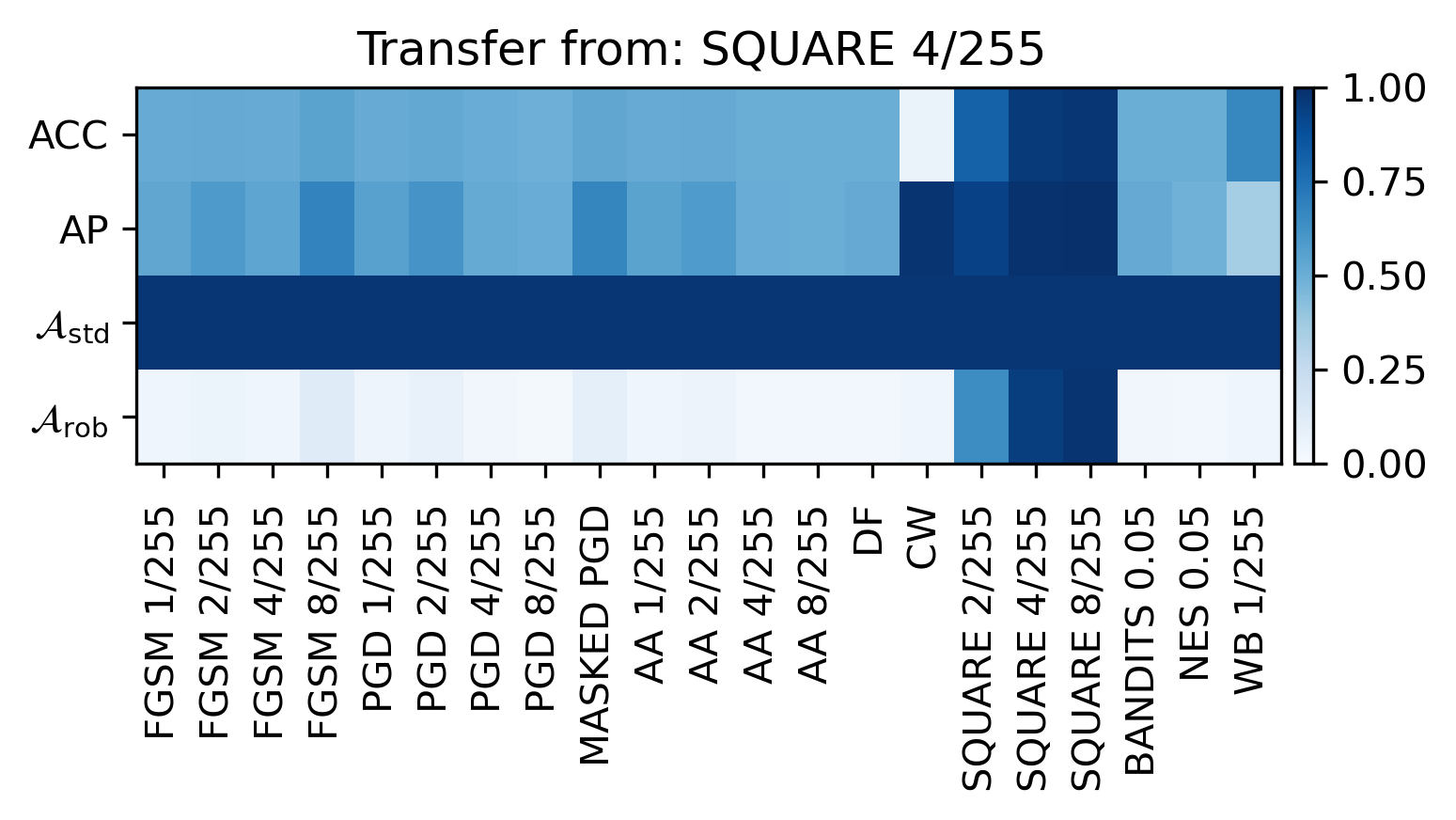}
        \caption{Transferability. From NES to the other attacks.}
        \label{fig:transfer_nes}
    \end{minipage}
\end{figure}

In this section, we plot the transfer capabilities from one dataset to the other datasets. 
For the white-box attacks, we denote FGSM \cref{fig:transfer_fgsm}, PGD \cref{fig:transfer_pgd}, Masked PGD \cref{fig:transfer_mPGD}, AA \cref{fig:transfer_aa},  DF \cref{fig:transfer_df}, and CW \cref{fig:transfer_cw}. 
DF  is a very strong attack but transfers well to other white-box attacks in \cref{fig:transfer_df}.
For the black-box attacks, we denote Square \cref{fig:transfer_square}, Bandits \cref{fig:transfer_bandits}, and NES \cref{fig:transfer_nes}.
In \cref{fig:transfer_nes}, we can observe that the perturbations transfer well to FGSM, PGD, and AA on the perturbation size of $\epsilon=8/255$.


